%% file: main.tex
\documentclass[10pt,twocolumn,letterpaper]{article}
\PassOptionsToPackage{table}{xcolor}
\usepackage[pagenumbers]{cvpr}
\input{preamble}

\usepackage{fontawesome}
\input{support/prompt_boxes}
\input{support/supplement_layout}
\definecolor{cvprblue}{rgb}{0.21,0.49,0.74}
\usepackage[pagebackref,breaklinks,colorlinks,allcolors=cvprblue]{hyperref}
\def\confName{CVPR}
\def\confYear{2027}
\title{\raisebox{-6pt}[0pt][0pt]{\includegraphics[height=26pt]{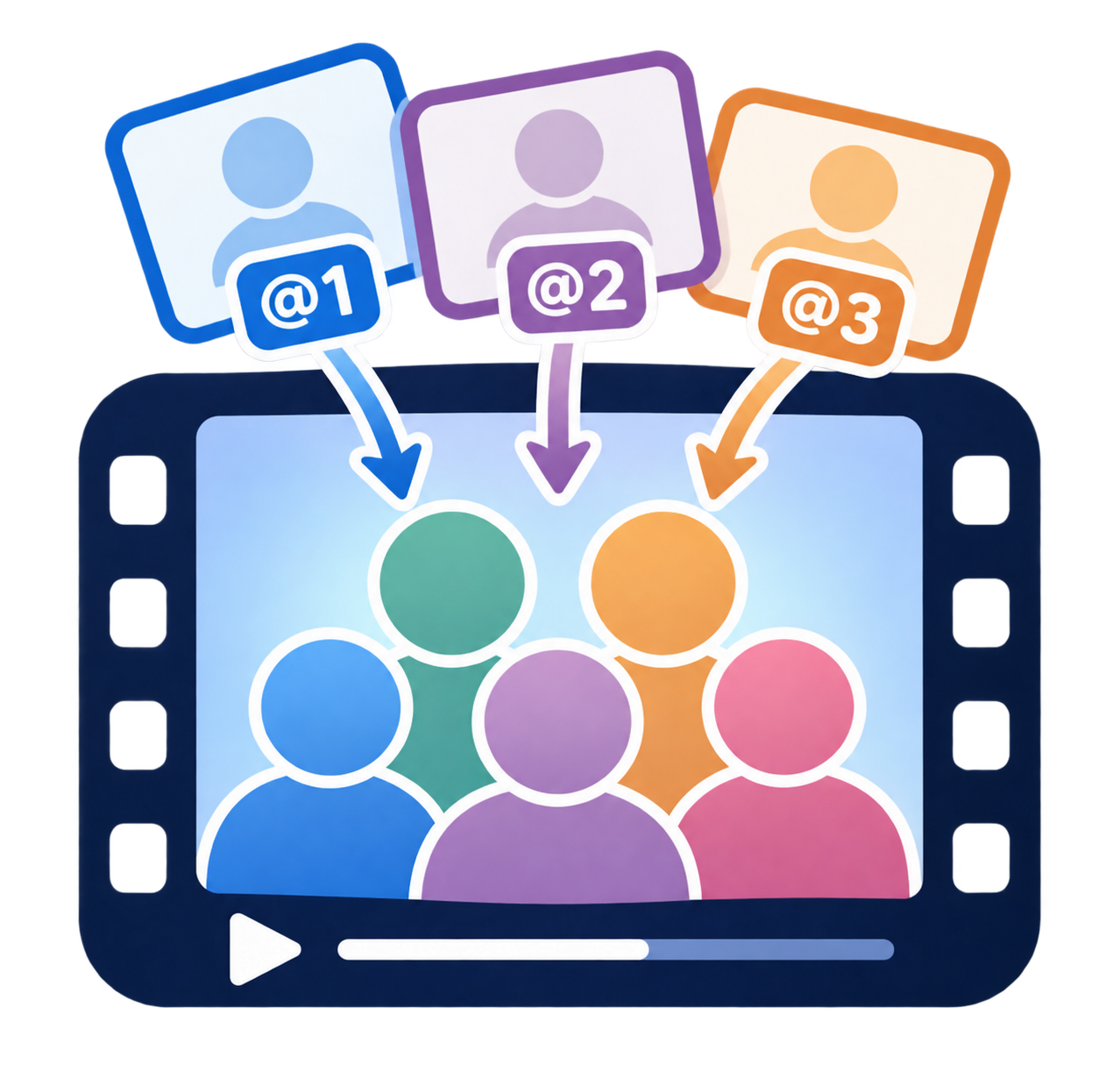}}\hspace{0.3em}%
GroupRef: Benchmarking and Improving Identity-Faithful Video Generation with Multi-Person References}
\author{
Xuan Wang$^{1}$\thanks{Equal contribution.} \qquad
Song Wang$^{1,2}$\footnotemark[1] \qquad
Bing Deng$^{2}$ \qquad
Yiru Zhao$^{2}$\thanks{Corresponding author.} \\[4pt]
$^{1}$Zhejiang University \qquad $^{2}$Alibaba Group \\[6pt]
\faGlobe~\textbf{Project Page:} \href{https://groupref.github.io/}{\texttt{https://GroupRef.github.io}}
}
\begin{document}
\maketitle

\input{sec/paper}

{\small
\bibliographystyle{ieeenat_fullname}
\bibliography{reference}
}

\clearpage

\appendix
\setcounter{figure}{0}
\setcounter{table}{0}
\setcounter{equation}{0}
\renewcommand{\thefigure}{S\arabic{figure}}
\renewcommand{\thetable}{S\arabic{table}}
\renewcommand{\theequation}{S\arabic{equation}}
\renewcommand{\theHfigure}{supp.\arabic{figure}}
\renewcommand{\theHtable}{supp.\arabic{table}}
\renewcommand{\theHequation}{supp.\arabic{equation}}

\makesupplementfrontmatter

\input{sec/appendix}

\end{document}

%% file: preamble.tex
\usepackage[T1]{fontenc}

\input{math_commands.tex}

\usepackage{url}
\usepackage{booktabs}
\usepackage{multirow}
\usepackage{xcolor}
\usepackage{graphicx}
\usepackage{tabularx}
\usepackage{framed}
\usepackage{placeins}

\DeclareMathSizes{7.5}{7}{5}{5}
\DeclareMathSizes{8.5}{8}{6}{5}

%% file: math_commands.tex
\usepackage{amsmath,amsfonts,bm}

\def\eqref#1{equation~\ref{#1}}

\def\1{\bm{1}}

\DeclareMathAlphabet{\mathsfit}{\encodingdefault}{\sfdefault}{m}{sl}
\SetMathAlphabet{\mathsfit}{bold}{\encodingdefault}{\sfdefault}{bx}{n}

%% file: support/prompt_boxes.tex
\usepackage{tcolorbox}
\tcbuselibrary{breakable}
\usepackage{fvextra}
\definecolor{promptBlue}{HTML}{356AA0}
\definecolor{promptTeal}{HTML}{287D78}
\newtcolorbox{evalpromptbox}[2][promptBlue]{
 breakable,colback=#1!3!white,colframe=#1!70!white,
 colbacktitle=#1!13!white,coltitle=#1!65!black,fonttitle=\sffamily\bfseries\small,
 title={#2},boxrule=0.45pt,arc=2pt,
 left=10pt,right=10pt,top=8pt,bottom=8pt,
 before skip=10pt,after skip=12pt,
 pad at break*=1mm,
 title after break={#2\space (continued)}
}

%% file: support/supplement_layout.tex
\usepackage{stfloats}
\makeatletter
\renewcommand{\l@section}[2]{%
  \addvspace{4pt}%
  \@dottedtocline{1}{0em}{1.5em}{\textbf{#1}}{\textbf{#2}}%
}
\renewcommand{\l@subsection}{\@dottedtocline{2}{1.5em}{2.4em}}
\newcommand{\supplementtitle}{\large The Supplementary Material of\\[8pt]
\LARGE\raisebox{-5pt}[0pt][0pt]{\includegraphics[height=25pt]{figures/groupref_icon.png}}\hspace{0.25em}%
GroupRef: Benchmarking and Improving Identity-Faithful Video Generation with Multi-Person References}
\let\supplement@maketitle\@maketitle
\newcommand{\makesupplementfrontmatter}{%
  \twocolumn[%
    \vskip .25in
    \begin{center}
      {\Large \bf \supplementtitle \par}
      \vspace*{16pt}
    \end{center}%
  ]%
}
\makeatother

%% file: sec/paper.tex
\begin{abstract}
Multi-person reference-to-video generation requires preserving individual identities while assigning actions to the corresponding people. Existing benchmarks often provide limited coverage of larger groups and allow textual appearance descriptions, making it difficult to distinguish whether models derive identity cues from text or visual references. To address these gaps, we introduce GroupRefBench, a benchmark featuring 3--10 distinct people per sample with appearance-free indexed instructions and an identity-grounded protocol for decoupled evaluation. Benchmarking 20 models on GroupRefBench reveals clear scaling bottlenecks in identity preservation and person-specific action following. To support model training, we also construct GroupRef100K, a large-scale dataset pairing diverse reference images with appearance-free indexed action instructions. Based on GroupRef100K, we train GroupRef with shared learnable anchors between image-index text tokens and corresponding reference-image tokens. Experiments on GroupRefBench and OpenS2V-Eval show that GroupRef achieves promising results in identity preservation and overall score, demonstrating the potential of GroupRef100K for training multi-person reference-to-video models.
\end{abstract}

\begin{table*}[t]
\centering
\caption{Comparison of reference-to-video benchmarks. Appearance denotes descriptions of faces, hairstyles, and reference clothing.}
\label{tab:benchmark-comparison}
\begingroup
\footnotesize
\setlength{\tabcolsep}{2.5pt}
\renewcommand{\arraystretch}{1.08}
\begin{tabular}{@{}lccccc@{}}
\toprule
\multirow{2}{*}{\textbf{Benchmark}}
& \multicolumn{3}{c}{\textbf{Input Protocol}}
& \multicolumn{2}{c}{\textbf{Evaluation Protocol}} \\
\cmidrule(lr){2-4}\cmidrule(lr){5-6}
& \textbf{\# Refs.}
& \textbf{Addressing}
& \textbf{Appearance}
& \textbf{Identity}
& \textbf{Action scoring} \\
\midrule
MSRVTT-Personalization~\citep{chen2025videoalchemist}
& 1--4 & Text & Allowed
& Visual similarity & Video--text similarity \\
OpenS2V-Eval~\citep{yuan2025opens2vnexus}
& 1--3 & Text & Allowed
& Visual similarity & Video--text similarity \\
UniVBench (R2V)~\citep{wei2026univbench}
& 1--3 & Text & Allowed
& VLM scoring & VLM scoring \\
IntelligentVBench (MI2V)~\citep{pan2026omniweaving}
& 1--4 & Indices & Allowed
& VLM scoring & VLM scoring \\
FashionVideoBench (MI2V)~\citep{wu2026loomvideo}
& Not reported & Indices & Allowed
& VLM scoring & VLM scoring \\
MultiRef-Compass~\citep{zhang2026multirefcompass}
& 3 & Indices & Allowed
& Visual similarity & VLM scoring \\
OmniVBench (Multi-content)~\citep{omnivbench2026}
& 2--5 & Indices & Allowed
& VLM scoring & VLM scoring \\
\midrule
\rowcolor{black!4}
\textbf{GroupRefBench}
& \textbf{3--10} & \textbf{Indices} & \textbf{Excluded}
& \textbf{Visual similarity + ID coverage}
& \textbf{ID matching $\rightarrow$ VLM scoring} \\
\bottomrule
\end{tabular}
\endgroup
\end{table*}

\section{Introduction}

Reference-to-video generation aims to synthesize videos conditioned on visual references and text instructions~\citep{liu2025phantom,wang2026refalign}.
In multi-person scenes, models need to preserve individual identities while accurately assigning actions to the corresponding people.
Recently, a growing trend is to use index markers such as \texttt{@Image k} to explicitly bind text instructions to specific reference people, removing the need for models to infer image--text correspondence~\citep{kling2026omni,byteplus2026seedance25,alibaba2026wan3,viduSubjectsReference}.
Alongside these advances, several benchmarks have been proposed to evaluate reference-to-video generation, as shown in Tab.~\ref{tab:benchmark-comparison}.
However, they remain limited in this multi-person setting in three key respects.
First, they include only a small number of references, leaving larger-group interactions unexplored.
Second, their prompts often allow textual appearance descriptions such as clothing, hairstyles, or facial features, which provide text shortcuts and obscure whether models rely on visual references or textual cues.
Third, existing evaluation protocols typically rely on holistic VLM scoring over full frames without prior person localization and identity matching, allowing videos to score well despite swapped identities or misassigned actions.

To address these limitations, we introduce \textbf{GroupRefBench}, a multi-person benchmark featuring 3--10 distinct individuals per sample.
Its prompts use image indices to assign actions to specific people while excluding appearance descriptions, removing textual appearance shortcuts and requiring models to rely on visual references.
Furthermore, the benchmark spans balanced group sizes with rich scene contexts and diverse multi-person interactions.
For evaluation, GroupRefBench introduces an identity-grounded protocol that decouples identity preservation, person-specific action following, group composition, and visual quality.
Specifically, it first tracks person instances in the generated video and matches them to the input references using facial and garment cues.
Based on these matches, we measure identity preservation across facial similarity, identity coverage, temporal consistency, and garment similarity, and verify whether each matched person performs the assigned action.
Additionally, group composition evaluates the overall person count including unmatched individuals, while visual quality is assessed independently.
By decoupling these dimensions, our protocol prevents high visual quality from obscuring missing identities or misassigned actions.

To facilitate model training, we further construct \textbf{GroupRef100K}, a large-scale dataset pairing target videos with diverse reference images and appearance-free indexed action instructions.
Our curation pipeline tracks individuals across frames to select clear per-person references, rewrites captions into indexed action instructions without appearance cues, and applies cross-shot pairing and reference editing to mitigate copy-paste artifacts.
On top of GroupRef100K, we train \textbf{GroupRef} on the RefAlign and MiniMax-H3 backbones by incorporating shared learnable anchors between image-index text tokens and their corresponding reference-image tokens.

We benchmark 20 open-source and commercial models on GroupRefBench.
Our evaluation reveals that although current models achieve high visual quality and roughly accurate person counts, they still struggle with identity preservation and person-specific action following as group size increases.
Moreover, experiments on both GroupRefBench and OpenS2V-Eval~\citep{yuan2025opens2vnexus} show that GroupRef consistently improves identity preservation and achieves competitive overall results over the corresponding backbones.

Our main contributions are summarized as follows:
\begin{itemize}
    \item We introduce \textbf{GroupRefBench}, a multi-person reference-to-video benchmark paired with an identity-grounded protocol for decoupled evaluation, and benchmark 20 models to reveal key scaling bottlenecks.
    \item We construct \textbf{GroupRef100K}, a large-scale training dataset that grounds multi-person references to appearance-free indexed action instructions, and train \textbf{GroupRef} models on it.
    \item Extensive experiments on GroupRefBench and OpenS2V-Eval show that GroupRef achieves promising results, particularly in identity preservation.
\end{itemize}

\begin{figure*}[t]
\centering
\includegraphics[width=\textwidth]{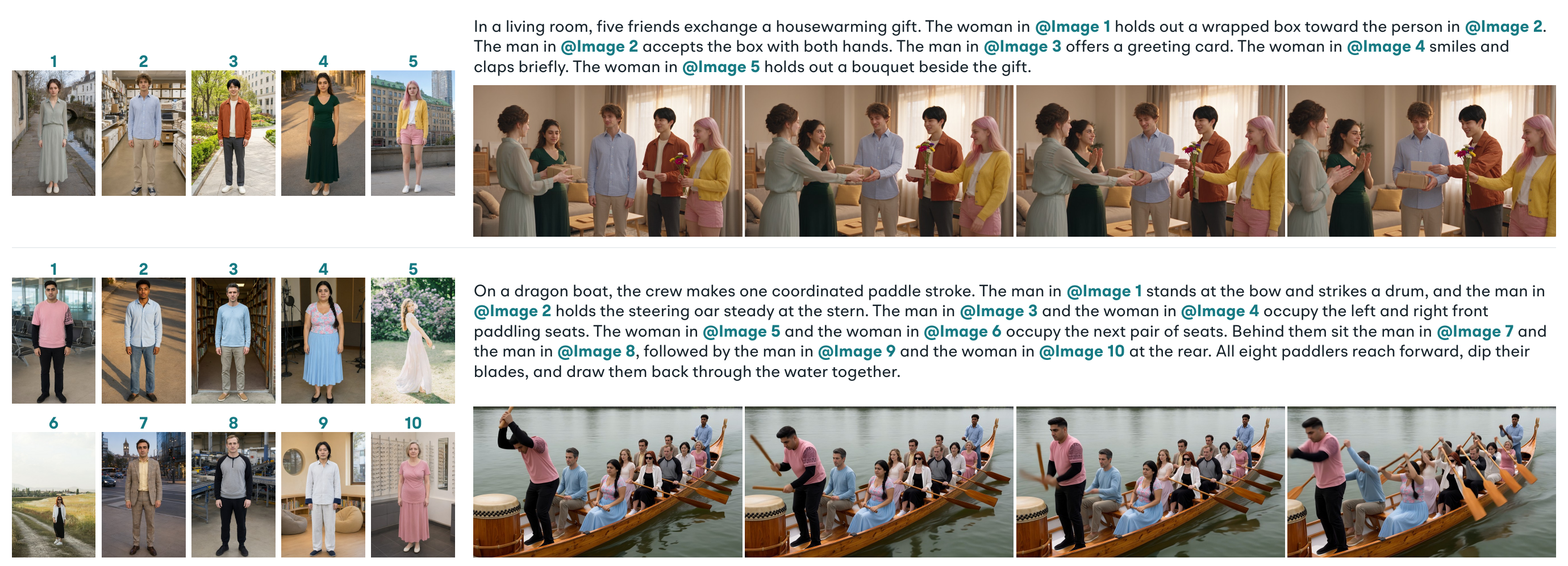}
\caption{Representative examples from GroupRefBench, pairing indexed reference portraits with an appearance-free \texttt{@Image k} prompt and sampled frames from an example generated video.}
\label{fig:bench-overview}
\end{figure*}

\section{Related Work}

\subsection{Reference-to-Video Benchmarks and Datasets}

Early reference-to-video benchmarks rely on text-addressed prompts and similarity metrics~\citep{chen2025videoalchemist,yuan2025opens2vnexus}, whereas subsequent works adopt VLM-based scoring under text~\citep{wei2026univbench} or index addressing~\citep{pan2026omniweaving,wu2026loomvideo}.
Recent benchmarks further examine fine-grained correspondence: MultiRef-Compass evaluates identity, attribute, and action binding~\citep{zhang2026multirefcompass}, and OmniVBench uses checklist-based VLM scoring to assess multi-content reference fidelity~\citep{omnivbench2026}.
However, as summarized in Tab.~\ref{tab:benchmark-comparison}, existing benchmarks cover at most five references, allow appearance descriptions, and score actions without explicit identity matching.
In contrast, GroupRefBench evaluates 3--10 individuals under appearance-free indexed prompts and grounds action scoring in prior person tracking and reference matching.

Training reference-to-video models requires reliable correspondences between video subjects and diverse reference images.
OpenS2V-Nexus pairs subject association with synthetic references~\citep{yuan2025opens2vnexus}, while Phantom-Data~\citep{chen2025phantomdata}, MuSS~\citep{zhang2026muss}, and Vera~\citep{xu2026vera} retrieve matching subjects or individuals across scenes and shots.
For text annotation, RefCaptioner links caption phrases to reference images~\citep{liu2026refcaptioner}, ReBind specifies reference sources in editing instructions~\citep{liu2026rebind}, and GroupVideo provides multi-person videos with captions for individual actions and interactions~\citep{song2026groupvideo}.
Building on these efforts, GroupRef100K pairs multi-person videos with appearance-free indexed prompts while diversifying reference images via cross-shot pairing and editing.

\subsection{Multi-Subject Reference-to-Video Generation}

Multi-subject reference-to-video methods condition video diffusion models on reference images to preserve subject appearance~\citep{liu2025phantom,fei2025skyreelsa2,jiang2025vace}, often using reference-text fusion to align visual features with prompt descriptions~\citep{chen2025videoalchemist,huang2025conceptmaster,hu2025polyvivid,li2025bindweave}.
To reduce identity confusion, Ingredients routes facial features to spatial video regions~\citep{fei2025ingredients}, MAGREF associates reference regions with prompt labels~\citep{deng2026magref}, and RefAlign enforces feature alignment~\citep{wang2026refalign}.
Person-focused methods further improve identity fidelity through explicit identity supervision~\citep{xu2026vera,song2026groupvideo} or VLM-guided reinforcement learning~\citep{pan2026idcrafter}.

Recent methods also organize multiple references within diffusion transformers through structured attention~\citep{xing2026lumosx} or specialized embeddings, including reference-specific positional encodings~\citep{zhang2025kaleido}, category embeddings~\citep{zhou2026aura}, and subject-shared embeddings~\citep{cai2026homie}.
For explicit text-reference binding, Movie Weaver combines anchored prompts with learnable concept embeddings on reference tokens~\citep{liang2025movieweaver}, and AlcheMinT shares index embeddings between reference tokens and separately encoded word tags~\citep{Girish_2026_CVPR}.
Following this direction, we equip existing backbones with Cross-Modal Identity Anchoring (CMIA), which shares learnable anchors between image-index tokens and their corresponding reference-image tokens, and train them on GroupRef100K to build our GroupRef baselines.

\section{GroupRefBench}
\label{sec:bench}

\subsection{Benchmark Construction}
\label{sec:bench-construction}

\begin{figure*}[t]
\centering
\includegraphics[width=\textwidth]{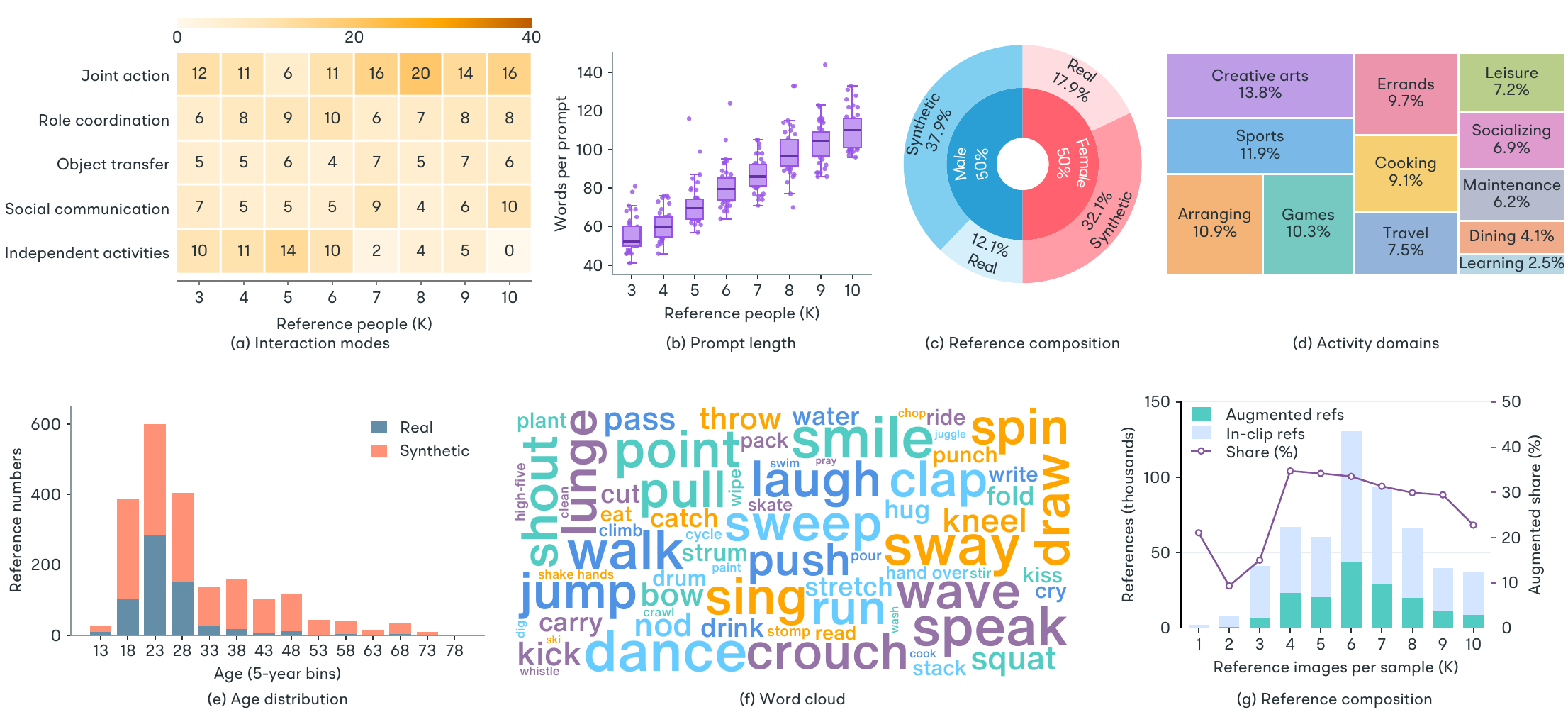}
\caption{Statistics of GroupRefBench (a--e) and GroupRef100K (f--g): (a) interaction modes across $K$, (b) prompt lengths, (c) gender and source breakdown, (d) activity domains, (e) age distribution, (f) frequent action verbs, and (g) reference counts and augmented shares.}
\label{fig:dataset-overview}
\end{figure*}

GroupRefBench contains 320 samples, with 40 samples for each group size $K \in \{3,\ldots,10\}$.
As illustrated in Fig.~\ref{fig:bench-overview}, each sample pairs an indexed prompt with $K$ distinct reference portraits.
These portraits are selected from a curated pool of 1,000 images comprising roughly 30\% real photographs from Pexels~\citep{pexels} and 70\% synthetic portraits generated by Qwen-Image-3.0-Pro~\citep{qwen2026qwenimage3}, with balanced gender representation and a broad age distribution, as shown in Fig.~\ref{fig:dataset-overview}(c, e).
We manually design seed scenarios across 12 everyday activity categories and five primary participation modes covering shared and individual actions, as summarized in Fig.~\ref{fig:dataset-overview}(a, d).
We then use Qwen3.8-Max~\citep{alibaba2026qwen38max} to expand these seeds into detailed prompts specifying scene context, objects, spatial relations, and temporal progression, resulting in prompt lengths that grow naturally with group size, as shown in Fig.~\ref{fig:dataset-overview}(b).
To avoid textual appearance shortcuts, the prompts assign actions through \texttt{@Image k} markers without any descriptions of physical appearance, hairstyle, or clothing, retaining only generic gender nouns and activity roles.
All samples are manually verified for identity distinctness, absence of appearance cues, instruction clarity, and scene plausibility.

\subsection{Evaluation Metrics}
\label{sec:metrics}

To evaluate performance on GroupRefBench, we design nine metrics across three dimensions: \textbf{identity preservation and group composition (IC)}, \textbf{visual quality (VQ)}, and \textbf{instruction following (IF)}.
Since person-level measurements require linking generated individuals to references, our pipeline begins with automated participant tracking and role matching to ground subsequent metrics.

\noindent\textbf{Participant Tracking and Role Matching.}
Given a generated video, we detect and segment individuals across frames, merging duplicate detections and fragmented tracklets via body appearance and temporal continuity while keeping co-occurring individuals separate.
Using video content alone without access to the target group size $K$ or reference images, we filter out incidental background pedestrians and people depicted in photographs or screens to retain only principal participants.
Empirically, the resulting participant counts closely match human annotations.
For each retained participant, we extract facial and garment features and apply the Hungarian algorithm~\citep{kuhn1955hungarian} for one-to-one matching against the $K$ references.
This assignment prioritizes pairs with valid features and maximizes weighted facial and garment similarity, renormalizing weights when one modality is occluded.
Without imposing a hard facial similarity threshold, this matching establishes candidate role assignments so that role-level action execution and facial identity fidelity can be diagnosed separately.

\noindent\textbf{Identity Preservation and Group Composition (IC).}
Based on the tracked participants and matched pairs, we compute five metrics:
(1) \textit{Face Similarity (Face)} computes the mean signed facial cosine similarity across matched pairs with visible faces;
(2) \textit{Identity Coverage (IDCov)} measures the fraction of the $K$ references whose matched participant meets a fixed facial similarity threshold;
(3) \textit{Count Accuracy (Count)} penalizes the relative error between the detected participant count and the target group size $K$;
(4) \textit{Temporal Identity Consistency (TempID)} evaluates within-track facial stability independently of references by computing the mean facial cosine similarity across temporally distant frame pairs on qualified tracks, averaged equally across eligible participants; and
(5) \textit{Garment Similarity (Garment)} measures visual similarity between each reference outfit and the matched participant's segmented clothing over time.
For \textit{Face}, \textit{TempID}, and \textit{Garment}, instances failing visibility or track-span criteria are excluded from the conditional mean.

\noindent\textbf{Visual Quality (VQ).}
Independent of role matching, we evaluate frame-level visual fidelity across sampled frames using two metrics:
(6) \textit{Image Quality (IQ)}, computed via MUSIQ~\citep{ke2021musiq}, and
(7) \textit{Aesthetic Quality (Aes)}, predicted by the LAION aesthetic regressor~\citep{schuhmann2022aesthetic}.

\noindent\textbf{Instruction Following (IF).}
For each sample, we construct a prompt-specific checklist covering participant actions and global scene requirements.
GPT-5.5~\citep{openai2026gpt55} scores each checklist item given 16 chronologically sampled frames annotated with the matched \texttt{@Image k} role labels, yielding two metrics:
(8) \textit{Role-Conditioned Action Following (Action)}, which evaluates whether each matched participant performs their assigned action with equal weight per requested role, and
(9) \textit{Other Instruction Following (Other)}, which evaluates scene context, objects, and spatial layout with equal weight per item.
Unjudgeable items receive no credit without redistributing their weights.

\noindent\textbf{Metric Aggregation.}
Each metric is averaged equally across valid videos, and the overall score (\textit{Total}) is computed as a fixed weighted sum of all nine metrics, with full formulations and human validation detailed in Appendix~\ref{app:metrics}.

\begin{figure*}[t]
\centering
\includegraphics[width=0.95\textwidth]{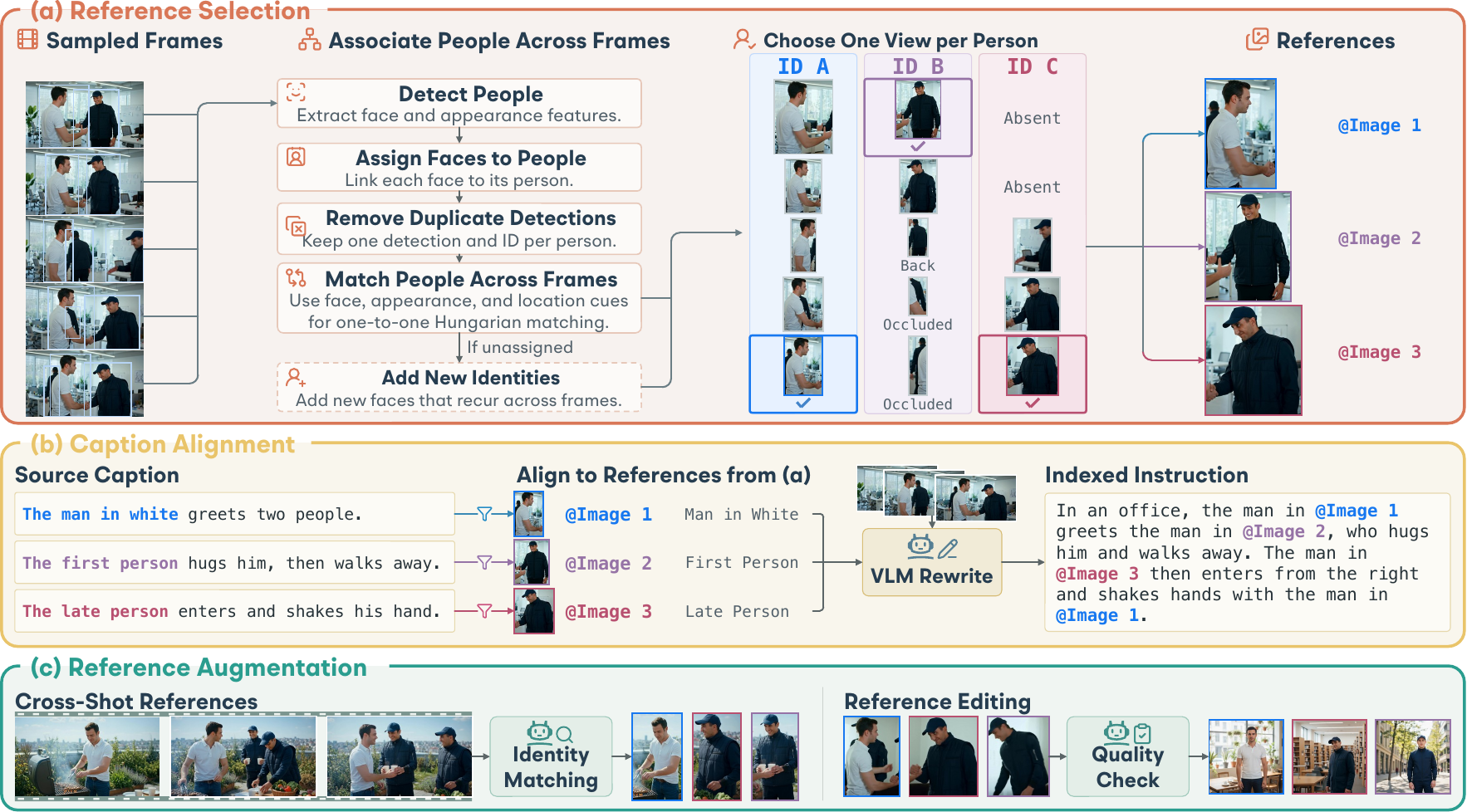}
\caption{Construction pipeline of GroupRef100K: (a) per-person reference selection, (b) appearance-free indexed caption alignment, and (c) reference augmentation via cross-shot pairing and background editing.}
\label{fig:pipeline}
\end{figure*}

\section{GroupRef100K}
\label{sec:dataset}

We curate \textbf{GroupRef100K} from 500,000 raw videos, yielding 100,000 samples that ground reference images to specific individuals and their actions in target videos.
Each sample pairs a target video and an appearance-free indexed action instruction with $K \in \{1,\ldots,10\}$ reference images, predominantly featuring multi-person interactions.
As illustrated in Fig.~\ref{fig:pipeline}, our data pipeline comprises three stages: reference selection, caption alignment, and reference augmentation, as detailed in Appendix~\ref{app:data}.
The resulting reference composition and action vocabulary are summarized in Fig.~\ref{fig:dataset-overview}(f, g).

\noindent\textbf{Reference Selection.}
Multiple participants rarely appear simultaneously with clear, unoccluded views in a single video frame.
To select an optimal view per person independently, we construct per-identity candidate pools from frames sampled at 4 fps, as shown in Fig.~\ref{fig:pipeline}(a).
We first detect bodies and faces in each frame, perform one-to-one face-to-person assignment, and suppress duplicate detections.
Hungarian matching then associates person detections across frames using facial embeddings, CLIP appearance features~\citep{radford2021clip}, and spatial cues, while clustering unassigned faces that persist over time into additional identities.
Finally, we select the highest-quality crop from each candidate pool based on facial frontality, body completeness, sharpness, and minimal occlusion, assigning each identity a unique \texttt{@Image k} index.

\noindent\textbf{Caption Alignment.}
In this stage, we align visual references with textual descriptions to convert source captions into appearance-free indexed action instructions, as shown in Fig.~\ref{fig:pipeline}(b).
A vision-language model (VLM) matches person descriptions in the source caption to the selected \texttt{@Image k} references, while supplementing unmentioned participants directly from sampled video frames.
Conditioned on this mapping and the video frames, the VLM rewrites the caption into a coherent instruction that replaces appearance descriptions with \texttt{@Image k} markers while preserving the original actions, interactions, spatial relations, and temporal order.

\noindent\textbf{Reference Augmentation.}
Training directly on reference crops extracted from target videos often induces copy-paste artifacts due to shared poses and backgrounds~\citep{zhang2025kaleido}.
We mitigate this issue through two complementary augmentation strategies while keeping the assigned indices, target videos, and instructions unchanged, as shown in Fig.~\ref{fig:pipeline}(c).
\textit{Cross-shot pairing} retrieves alternative views of the same identity from other shots via mutual nearest-neighbor search, filtering candidates using facial similarity, appearance consistency, and VLM verification.
\textit{Reference editing} synthesizes diverse backgrounds to produce standalone portraits while preserving facial identity, hairstyle, and clothing.
In both cases, an augmented candidate replaces the original reference only if it passes quality and identity-consistency checks; otherwise, the original crop is retained.

\section{GroupRef}
\label{sec:method}

\begin{figure*}[t]
\centering
\includegraphics[width=0.88\textwidth]{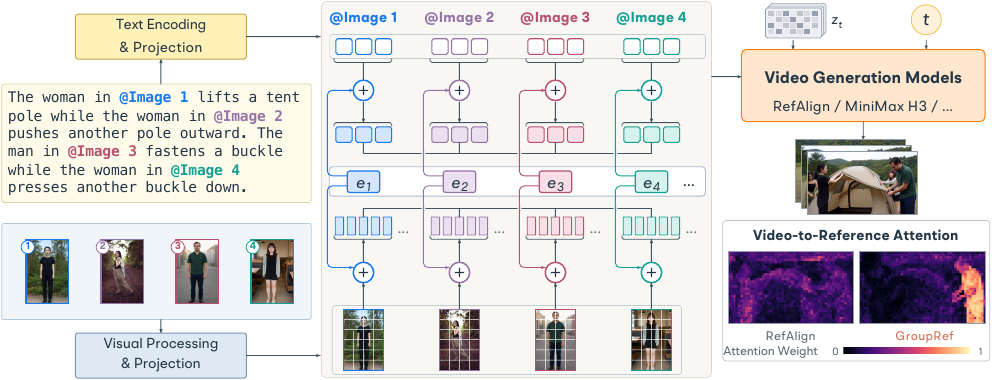}
\caption{Overview of GroupRef with Cross-Modal Identity Anchoring (CMIA), which adds a shared learnable anchor $e_k$ to both the text tokens of \texttt{@Image k} and the visual tokens of reference image $I_k$.}
\label{fig:model}
\end{figure*}

For indexed reference-to-video generation, we develop GroupRef, which incorporates a lightweight Cross-Modal Identity Anchoring (CMIA) mechanism to explicitly bind reference images with their textual mentions, as shown in Fig.~\ref{fig:model}.
Specifically, given a text prompt $c$ and $K$ reference images $\{I_k\}_{k=1}^{K}$, the text and visual projection modules first extract text features $\{c_i\}\subset\mathbb{R}^{d}$ and visual features $\{v_{k,p}\}\subset\mathbb{R}^{d}$, where $i$ indexes prompt tokens and $p$ indexes visual tokens within $I_k$.
CMIA maintains a learnable anchor table $E\in\mathbb{R}^{K_{\max}\times d}$ for up to $K_{\max}$ references, adding the $k$-th index anchor $e_k\in\mathbb{R}^{d}$ to both the visual tokens of $I_k$ and the text tokens of \texttt{@Image k}:
\begin{equation}
\begin{aligned}
\widetilde{c}_i &= c_i+\sum_{k=1}^{K}m_{ik}e_k,\\
\widetilde{v}_{k,p} &= v_{k,p}+e_k,
\end{aligned}
\label{eq:cmia}
\end{equation}
where $m_{ik}=1$ if token $i$ belongs to \texttt{@Image k} and $0$ otherwise.
Repeated mentions of \texttt{@Image k} share the same $e_k$, while all other prompt tokens remain unchanged.
The anchored features then condition the video generation model on noisy latents $z_t$ at each denoising timestep $t$, guiding video-to-reference attention toward the corresponding person.

We instantiate GroupRef on the RefAlign and MiniMax-H3 backbones and train them on GroupRef100K.
For each backbone, we apply LoRA~\citep{hu2022lora} and jointly optimize an independent anchor table initialized with $e_k\sim\mathcal{N}(0,\sigma^2 I_d)$ under its native training objective.
During training, we also randomly shuffle the reference images and synchronously update the corresponding \texttt{@Image k} indices in the prompt, encouraging each anchor to learn index correspondence instead of memorizing specific identities.
After training, GroupRef exhibits a clearer attention preference for the corresponding reference images than the original backbone under identical noised video inputs, as illustrated in the bottom-right panel of Fig.~\ref{fig:model}.
Training details and further model analyses are provided in Appendix~\ref{app:model-analyses}.

\section{Experiments}
\label{sec:experiments}

\subsection{Evaluation on GroupRefBench}
\label{sec:bench-results}

\input{tables/grouprefbench}

We evaluate 20 open-source and commercial models on GroupRefBench and report the quantitative results in Tab.~\ref{tab:grouprefbench-results}, where $N$ denotes the number of returned videos.
Because interface limits and prompt filtering in several commercial models lead to unequal returned counts, we also report a shared-subset evaluation in Appendix~\ref{app:k-coverage}.

\begin{figure}[t]
\centering
\includegraphics[width=\columnwidth]{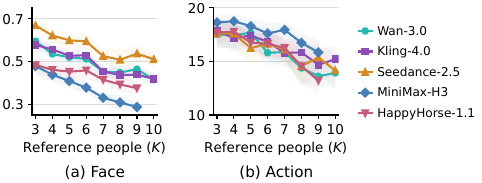}
\caption{Performance trends across reference people $K$.}
\label{fig:commercial-scaling}
\end{figure}

Commercial models achieve the highest overall scores on GroupRefBench.
Although open-source baselines reach comparable visual quality and count accuracy, they trail commercial models in identity preservation and person-specific action following, indicating that generating roughly accurate headcounts does not guarantee faithful identity binding.
Evaluating across group size $K$ further reveals a clear multi-person bottleneck.
As $K$ increases, open-source baselines show clear declines in participant count, identity preservation, and action following, whereas commercial models maintain stable visual quality and broad identity coverage while still declining in facial similarity, count accuracy, and action scores.
This scaling trend is illustrated in Fig.~\ref{fig:commercial-scaling}, where five commercial models exhibit consistent downward trends in both Face and Action scores as $K$ grows.

Across different backbones, GroupRef improves overall performance, with consistent gains in identity preservation and action following.
Even on a strong backbone such as MiniMax-H3-33B, GroupRef yields consistent improvements, supporting the effectiveness of GroupRef100K and our anchoring design.

\subsection{Evaluation on OpenS2V-Eval}
\label{sec:opens2v}

\input{tables/opens2v}

\begin{figure*}[t]
    \centering
    \includegraphics[width=0.9\textwidth]{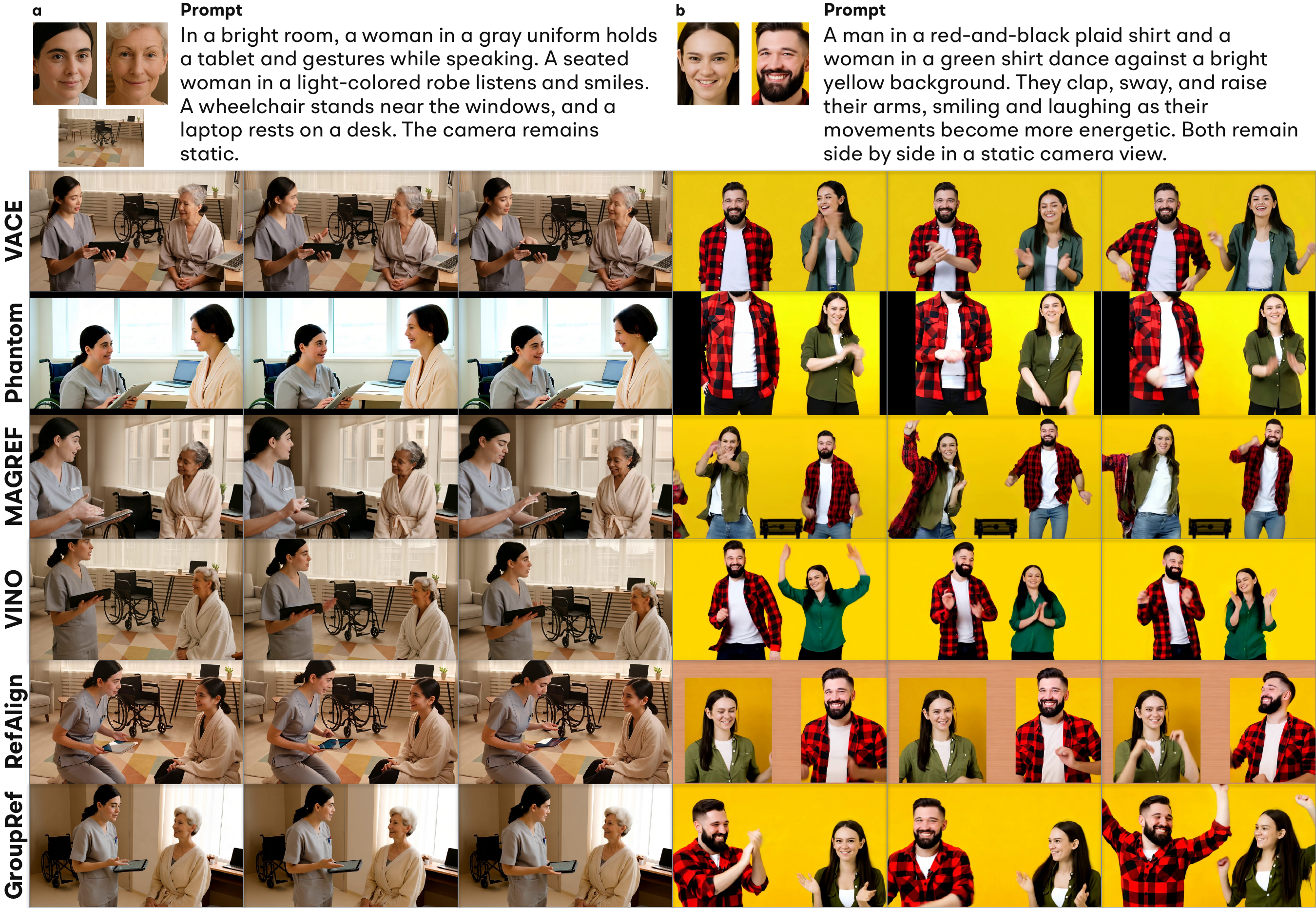}
    \caption{Qualitative comparisons on OpenS2V-Eval.}
    \label{fig:case}
\end{figure*}

To further assess generalization, we evaluate GroupRef on OpenS2V-Eval~\citep{yuan2025opens2vnexus}, a public benchmark containing 180 test samples.
As shown in Tab.~\ref{tab:opens2v-results}, GroupRef achieves competitive overall performance among the compared methods, particularly in facial similarity.
Fig.~\ref{fig:case} presents qualitative comparisons, showing that GroupRef preserves each subject's facial appearance alongside natural motion and coherent scene layouts.

\input{tables/ablation}

\subsection{Ablation Study}
\label{sec:ablation}
\label{sec:attention-analysis}

We conduct ablations with GroupRef-1.3B on GroupRefBench to evaluate the contributions of GroupRef100K and CMIA.
Specifically, we compare the full model against five variants:
(a) \emph{w/o CMIA}, removing all anchors;
(b) \emph{w/o image anchors} and (c) \emph{w/o text anchors}, applying anchors only to \texttt{@Image k} text tokens or reference-image tokens, respectively;
(d) \emph{w/ separate anchors}, using independent learnable anchors for the two modalities; and
(e) \emph{w/ fixed hash anchors}, replacing learnable anchors with fixed vectors derived from SHA-256 hashes of the reference images.

As shown in Tab.~\ref{tab:groupref-ablation}, variant (a) clearly improves over the original RefAlign-1.3B backbone in Tab.~\ref{tab:grouprefbench-results}, demonstrating the value of training on GroupRef100K.
Furthermore, the full GroupRef-1.3B outperforms single-modality, unshared, and hash-based variants to achieve the best overall performance, confirming the benefit of shared learnable index anchoring in CMIA.

\section{Conclusion}
In this paper, we introduced GroupRefBench, a multi-person reference-to-video benchmark pairing visual references of 3--10 distinct individuals with appearance-free indexed instructions and an identity-grounded evaluation protocol.
Benchmarking 20 open-source and commercial models reveals clear scaling bottlenecks in multi-person identity preservation and person-specific action following that visual quality and count metrics alone do not capture.
To support model training for this task, we constructed GroupRef100K with appearance-free indexed action instructions and trained GroupRef using shared learnable index anchors.
Experiments on both GroupRefBench and OpenS2V-Eval show that GroupRef achieves promising results, particularly in identity preservation, demonstrating the value of our dataset and training framework for multi-person reference-to-video generation.

%% file: tables/grouprefbench.tex
\begin{table*}[!t]
\centering
\caption{Quantitative results on GroupRefBench.}
\label{tab:grouprefbench-results}
\begingroup
\fontsize{8.3}{10.2}\selectfont
\setlength{\tabcolsep}{3.0pt}
\renewcommand{\arraystretch}{1.05}
\begin{tabular}{l*{11}{c}}
\toprule
& & & \multicolumn{5}{c}{\textbf{IC}} & \multicolumn{2}{c}{\textbf{VQ}} & \multicolumn{2}{c}{\textbf{IF}} \\
\cmidrule(lr){4-8}\cmidrule(lr){9-10}\cmidrule(lr){11-12}
\multirow{-2}{*}{\textbf{Method}} &
\multirow{-2}{*}{\textbf{$N$}} &
\multirow{-2}{*}{\textbf{Total}} &
\textbf{Face} & \textbf{IDCov} & \textbf{Count} &
\textbf{TempID} & \textbf{Garment} &
\textbf{IQ} & \textbf{Aes} &
\textbf{Action} & \textbf{Other} \\
\midrule
\multicolumn{12}{l}{\textit{Commercial models}} \\
Seedance-2.5~\citep{byteplus2026seedance25} & 259 & 74.25 & 0.573 & 0.946 & 0.930 & 0.715 & 0.721 & 64.82 & 5.27 & 16.11 & 8.71 \\
MiniMax-H3~\citep{minimax2026h3} & 279 & 73.11 & 0.375 & 0.945 & 0.982 & 0.594 & 0.780 & 74.71 & 5.50 & 17.68 & 9.39 \\
HappyHorse-1.1~\citep{alibaba2026happyhorse11} & 280 & 72.60 & 0.433 & 0.953 & 0.958 & 0.621 & 0.795 & 75.35 & 5.45 & 16.12 & 9.09 \\
Kling-4.0~\citep{kling2026v4} & 320 & 73.21 & 0.491 & 0.933 & 0.914 & 0.648 & 0.753 & 74.93 & 5.45 & 16.34 & 8.91 \\
Wan-3.0~\citep{alibaba2026wan3} & 320 & 73.35 & 0.492 & 0.964 & 0.945 & 0.663 & 0.835 & 74.45 & 5.35 & 15.77 & 8.57 \\
\midrule
\multicolumn{12}{l}{\textit{Open-source models}} \\
BindWeave-14B~\citep{li2025bindweave} & 320 & 24.37 & 0.003 & 0.015 & 0.627 & 0.429 & 0.456 & 48.30 & 4.26 & 0.88 & 3.18 \\
LoomVideo-5B~\citep{wu2026loomvideo} & 320 & 39.00 & 0.003 & 0.051 & 0.805 & 0.530 & 0.659 & 65.84 & 5.00 & 5.74 & 6.83 \\
MAGREF-14B~\citep{deng2026magref} & 320 & 38.90 & 0.018 & 0.089 & 0.811 & 0.511 & 0.686 & 71.45 & 5.21 & 4.64 & 6.56 \\
OmniWeaving-8.3B~\citep{pan2026omniweaving} & 320 & 39.13 & 0.017 & 0.096 & 0.728 & 0.453 & 0.652 & 70.07 & 4.92 & 7.13 & 6.19 \\
VACE-14B~\citep{jiang2025vace} & 320 & 40.65 & 0.011 & 0.081 & 0.863 & 0.487 & 0.663 & 74.73 & 5.33 & 5.33 & 7.31 \\
Kaleido-14B~\citep{zhang2025kaleido} & 320 & 41.02 & 0.064 & 0.207 & 0.779 & 0.574 & 0.705 & 63.66 & 5.02 & 4.45 & 7.26 \\
RefAlign-1.3B~\citep{wang2026refalign} & 320 & 40.73 & 0.060 & 0.181 & 0.785 & 0.599 & 0.724 & 75.38 & 5.23 & 3.45 & 6.78 \\
RefAlign-14B~\citep{wang2026refalign} & 320 & 44.71 & 0.104 & 0.302 & 0.846 & 0.509 & 0.737 & 76.20 & 5.41 & 4.76 & 7.49 \\
UniVideo-13B~\citep{wei2025univideo} & 320 & 45.79 & 0.204 & 0.447 & 0.734 & 0.641 & 0.784 & 67.91 & 5.15 & 3.81 & 6.65 \\
VINO-13B~\citep{chen2026vino} & 320 & 46.60 & 0.198 & 0.427 & 0.778 & 0.618 & 0.737 & 69.68 & 5.20 & 4.82 & 6.60 \\
Phantom-14B~\citep{liu2025phantom} & 320 & 47.09 & 0.192 & 0.369 & 0.821 & 0.633 & 0.740 & 74.06 & 5.33 & 4.23 & 7.27 \\
MiniMax-H3-33B~\citep{minimax2026h3} & 320 & 63.84 & 0.411 & 0.880 & 0.913 & 0.563 & 0.725 & 73.78 & 5.38 & 11.10 & 8.01 \\
\midrule
\rowcolor{black!4}
\textbf{GroupRef-1.3B} & 320 & 47.10 & 0.192 & 0.465 & 0.817 & 0.610 & 0.836 & 71.20 & 5.29 & 3.93 & 6.77 \\
\rowcolor{black!4}
\textbf{GroupRef-14B} & 320 & 50.66 & 0.232 & 0.601 & 0.860 & 0.641 & 0.763 & 69.69 & 5.08 & 5.34 & 6.81 \\
\rowcolor{black!4}
\textbf{GroupRef-33B} & 320 & 64.39 & 0.439 & 0.895 & 0.892 & 0.555 & 0.726 & 73.49 & 5.35 & 11.23 & 8.07 \\
\bottomrule
\end{tabular}
\endgroup

\end{table*}

%% file: tables/opens2v.tex
\begin{table*}[!t]
\centering
\caption{Quantitative results on OpenS2V-Eval.}
\label{tab:opens2v-results}
\begingroup
\fontsize{8.5}{10.2}\selectfont
\setlength{\tabcolsep}{3.5pt}
\renewcommand{\arraystretch}{1.05}
\begin{tabular}{l*{8}{c}}
\toprule
\textbf{Method}
& \textbf{Overall}
& \textbf{Aes}
& \textbf{Smooth}
& \textbf{Amp}
& \textbf{FaceSim}
& \textbf{GME}
& \textbf{Nexus}
& \textbf{Natural} \\
\midrule
\multicolumn{9}{l}{\textit{Commercial models}} \\
Kling-1.6~\citep{kling2025official}
& 56.23\% & 44.59\% & 86.93\% & 41.60\%
& 40.10\% & 66.20\% & 45.89\% & 74.59\% \\
Vidu-2.0~\citep{bao2024vidu}
& 51.95\% & 41.48\% & 90.45\% & 13.52\%
& 35.11\% & 67.57\% & 43.37\% & 65.88\% \\
Pika-2.1~\citep{pika2025official}
& 51.88\% & 46.88\% & 87.06\% & 24.71\%
& 30.38\% & 69.19\% & 45.40\% & 63.32\% \\
\midrule
\multicolumn{9}{l}{\textit{Research models}} \\
Saber~\citep{zhou2025saber}
& 57.91\% & 42.42\% & 96.12\% & 21.12\%
& 49.89\% & 67.50\% & 47.22\% & 72.55\% \\
VACE-1.3B~\citep{jiang2025vace}
& 49.89\% & 48.24\% & 97.20\% & 18.83\%
& 20.57\% & 71.26\% & 37.91\% & 65.46\% \\
BindWeave~\citep{li2025bindweave}
& 57.61\% & 45.55\% & 95.90\% & 13.91\%
& 53.71\% & 67.79\% & 46.84\% & 66.85\% \\
VINO~\citep{chen2026vino}
& 57.85\% & 45.92\% & 94.73\% & 12.30\%
& 52.00\% & 69.69\% & 42.67\% & 71.99\% \\
VACE-14B~\citep{jiang2025vace}
& 57.55\% & 47.21\% & 94.97\% & 15.02\%
& 55.09\% & 67.27\% & 44.08\% & 67.04\% \\
Phantom-14B~\citep{liu2025phantom}
& 56.77\% & 46.39\% & 96.31\% & 33.42\%
& 51.46\% & 70.65\% & 37.43\% & 69.35\% \\
MAGREF-480P~\citep{deng2026magref}
& 52.51\% & 45.02\% & 93.17\% & 21.81\%
& 30.83\% & 70.47\% & 43.04\% & 66.90\% \\
SkyReels-A2-P14B~\citep{fei2025skyreelsa2}
& 52.25\% & 39.41\% & 87.93\% & 25.60\%
& 45.95\% & 64.54\% & 43.75\% & 60.32\% \\
RefAlign-1.3B~\citep{wang2026refalign}
& 56.30\% & 42.96\% & 94.74\% & 20.74\%
& 53.06\% & 66.85\% & 43.97\% & 66.25\% \\
RefAlign-14B~\citep{wang2026refalign}
& 60.42\% & 46.84\% & 97.61\% & 22.48\%
& 55.23\% & 68.32\% & 48.52\% & 73.63\% \\
\midrule
\rowcolor{black!4}
\textbf{GroupRef-1.3B}
& 58.11\% & 43.05\% & 96.30\% & 20.40\%
& 60.61\% & 66.65\% & 43.86\% & 67.28\% \\
\rowcolor{black!4}
\textbf{GroupRef-14B}
& 60.99\% & 48.04\% & 97.94\% & 21.16\%
& 62.44\% & 68.74\% & 45.73\% & 71.33\% \\
\bottomrule
\end{tabular}
\endgroup
\end{table*}

%% file: tables/ablation.tex
\begin{table*}[!t]
\centering
\caption{Ablation study of GroupRef-1.3B on GroupRefBench.}
\label{tab:groupref-ablation}
\begingroup
\fontsize{8.3}{10.2}\selectfont
\setlength{\tabcolsep}{3.0pt}
\renewcommand{\arraystretch}{1.05}
\begin{tabular}{l*{11}{c}}
\toprule
& & & \multicolumn{5}{c}{\textbf{IC}} & \multicolumn{2}{c}{\textbf{VQ}} & \multicolumn{2}{c}{\textbf{IF}} \\
\cmidrule(lr){4-8}\cmidrule(lr){9-10}\cmidrule(lr){11-12}
\multirow{-2}{*}{\textbf{Method}} &
\multirow{-2}{*}{\textbf{$N$}} &
\multirow{-2}{*}{\textbf{Total}} &
\textbf{Face} & \textbf{IDCov} & \textbf{Count} &
\textbf{TempID} & \textbf{Garment} &
\textbf{IQ} & \textbf{Aes} &
\textbf{Action} & \textbf{Other} \\
\midrule
\textbf{GroupRef-1.3B} & 320 & \textbf{47.10} & \textbf{0.192} & \textbf{0.465} & 0.817 & \textbf{0.610} & \textbf{0.836} & 71.20 & \textbf{5.29} & \textbf{3.93} & \textbf{6.77} \\
\quad (a) w/o CMIA & 320 & 45.34 & 0.171 & 0.435 & 0.824 & 0.589 & 0.814 & 70.83 & 5.26 & 3.35 & 6.58 \\
\quad (b) w/o image anchors & 320 & 44.86 & 0.168 & 0.443 & 0.830 & 0.517 & 0.826 & 71.47 & 5.27 & 3.38 & 6.60 \\
\quad (c) w/o text anchors & 320 & 45.08 & 0.155 & 0.438 & \textbf{0.843} & 0.510 & 0.804 & \textbf{71.86} & 5.27 & 3.81 & 6.72 \\
\quad (d) w/ separate anchors & 320 & 43.78 & 0.148 & 0.389 & 0.831 & 0.504 & 0.802 & 71.52 & 5.25 & 3.61 & 6.43 \\
\quad (e) w/ fixed hash anchors & 320 & 45.45 & 0.180 & 0.457 & 0.819 & 0.580 & 0.814 & 71.25 & 5.27 & 3.21 & 6.55 \\
\bottomrule
\end{tabular}
\endgroup
\end{table*}

%% file: sec/appendix.tex
\section{Extended Benchmark Analyses}

We complement the main GroupRefBench experiments in Sec.~\ref{sec:bench-results} with per-group-size scaling trends (Appendix~\ref{app:by-k}), a controlled shared-subset evaluation (Appendix~\ref{app:k-coverage}), action coverage decompositions (Appendix~\ref{app:action-analysis}), and multi-reference conditioning protocols across all 20 evaluated models (Appendix~\ref{app:reference-adaptation}).

\subsection{Group-Size Scaling}
\label{app:by-k}
\label{app:k-trends}

\begin{figure*}[!b]
\centering
\includegraphics[width=\textwidth]{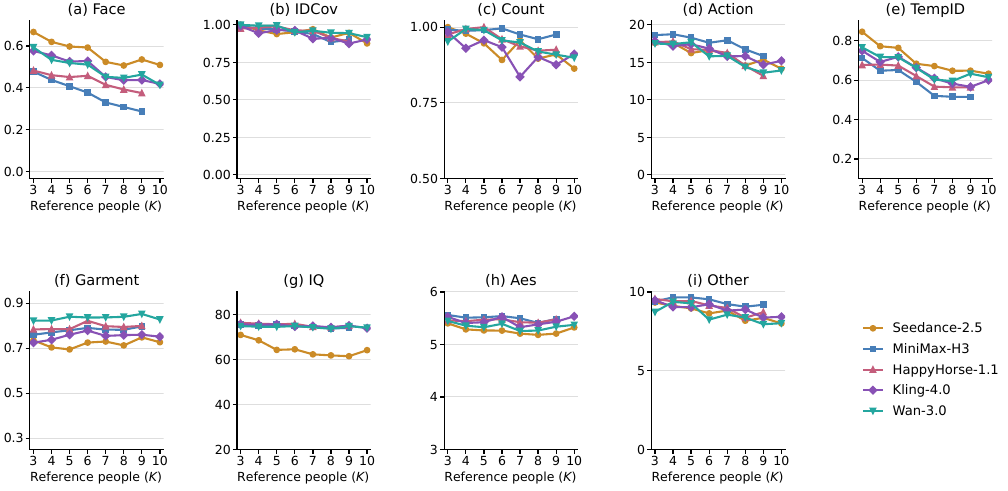}
\caption{Scaling trends across $K$ for commercial models.}
\label{fig:k-core-commercial}
\end{figure*}

\begin{figure*}[!tp]
\centering
\includegraphics[width=\textwidth]{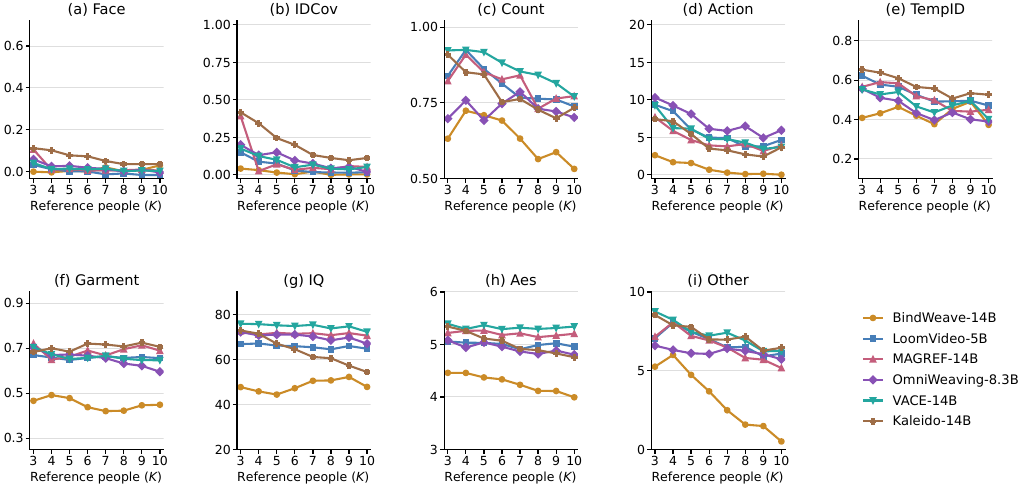}
\caption{Scaling trends across $K$ for open-source baselines, cohort I.}
\label{fig:k-core-open-a}
\end{figure*}

\begin{figure*}[!tp]
\centering
\includegraphics[width=\textwidth]{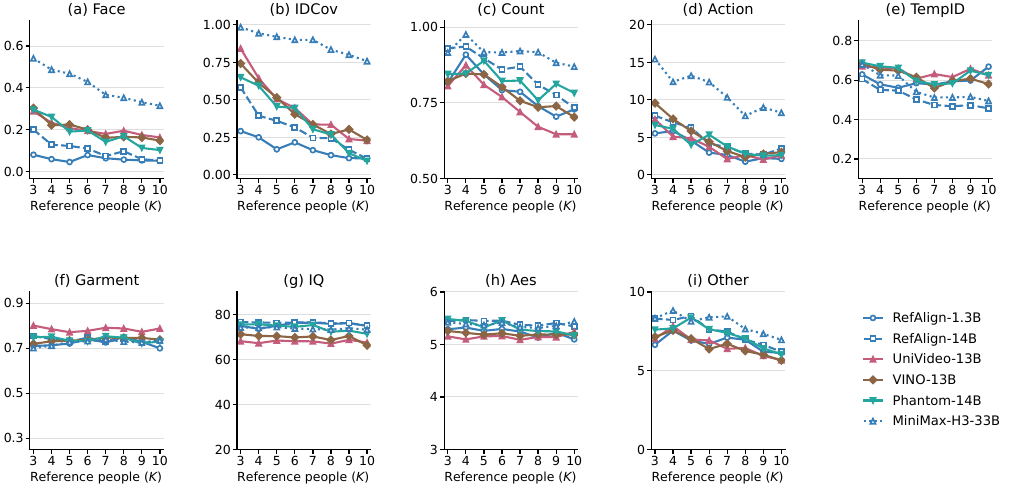}
\caption{Scaling trends across $K$ for open-source baselines, cohort II.}
\label{fig:k-core-open-b}
\end{figure*}

\begin{figure*}[!tp]
\centering
\includegraphics[width=\textwidth]{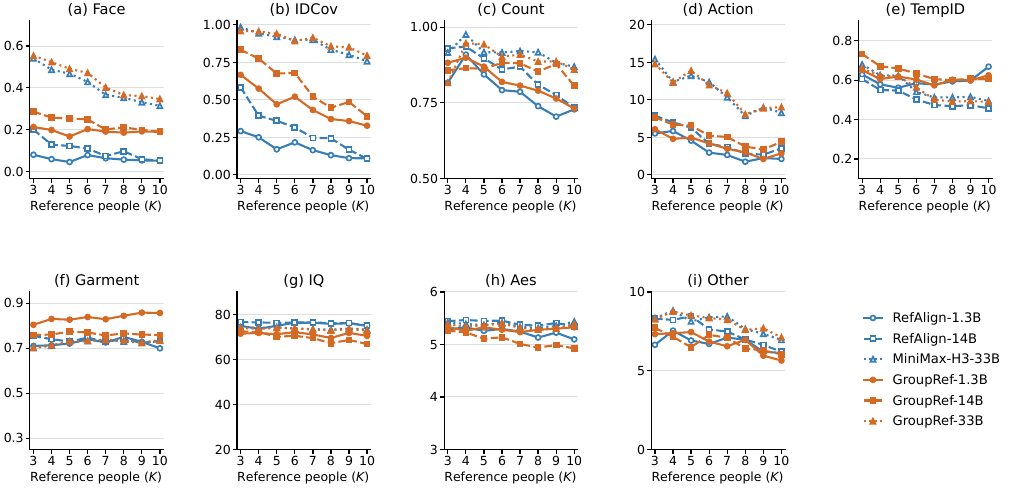}
\caption{Scaling trends across $K$ for GroupRef and its base backbones.}
\label{fig:k-core-backbones}
\end{figure*}

Scaling multi-person video generation requires preserving individual identities and assigned actions as the group size $K$ grows. As shown in Figs.~\ref{fig:k-core-commercial}--\ref{fig:k-core-backbones}, visual quality (\textbf{IQ}, \textbf{Aes}) and garment similarity (\textbf{Garment}) remain nearly constant across $K \in \{3, \ldots, 10\}$ for most of the 20 evaluated models. By contrast, identity preservation (\textbf{Face}, \textbf{IDCov}, \textbf{TempID}), count accuracy (\textbf{Count}), and action following (\textbf{Action}) generally decline as group size increases, indicating that current video models maintain overall visual quality and clothing colors even as identity binding and role execution degrade in larger groups.

Comparing scaling curves across model families reveals three clear trends:
\begin{itemize}
    \item \textbf{Commercial models.} As shown in Fig.~\ref{fig:k-core-commercial}, commercial systems, where MiniMax-H3 and HappyHorse-1.1 stop at $K=9$ due to 9-image API limits, maintain high identity coverage even in larger groups. Instead of omitting people, they exhibit a gradual drop in facial similarity and temporal consistency as each face occupies fewer pixels.
    \item \textbf{Open-source baselines.} Dedicated reference-to-video models in Fig.~\ref{fig:k-core-open-b}, including UniVideo-13B, VINO-13B, and Phantom-14B, start with strong identity coverage at $K=3$ but drop rapidly beyond $K=5$, accompanied by lower count and action scores. On the other hand, general multi-condition models in Fig.~\ref{fig:k-core-open-a} show very low facial similarity even at $K=3$; their flat \textbf{Face} curves simply reflect weak initial face binding and should be read together with \textbf{IDCov} and \textbf{Count}.
    \item \textbf{Identity vs.\ action scaling.} Comparing base backbones with their GroupRef counterparts in Fig.~\ref{fig:k-core-backbones} shows that multi-person training and index anchoring consistently lift \textbf{Face} and \textbf{IDCov} across group sizes, while overall \textbf{Action} levels remain strongly influenced by pretrained backbone scale.
\end{itemize}

\subsection{Shared-Subset Evaluation}
\label{app:k-coverage}

To separate model capability from API filtering and reference-count limits, we complement Tab.~\ref{tab:grouprefbench-results} with an evaluation on a shared video subset. Because MiniMax-H3 and HappyHorse-1.1 accept at most 9 reference images, omitting all 40 ten-person prompts, and Seedance-2.5 filters certain prompts across group sizes, Tab.~\ref{tab:k-common} evaluates all 20 models on the common intersection of $N=229$ scenes ($K \le 9$). Since $K=10$ is the most challenging setting, averaging over all returned videos slightly penalizes models that complete all 320 prompts: on the shared $N=229$ subset, Kling-4.0 and Wan-3.0 move slightly ahead of MiniMax-H3 and HappyHorse-1.1, while overall model tiers remain consistent.

The matched comparison in Tab.~\ref{tab:k-common} highlights four general patterns:
\begin{itemize}
    \item \textbf{Visual quality vs.\ identity preservation.} Visual quality saturates early across model scales: several open-source baselines, such as VACE-14B and RefAlign-14B, match or surpass commercial models in \textbf{IQ} and \textbf{Aes}, yet remain much lower in \textbf{IDCov} and \textbf{Action}.
    \item \textbf{Person count and clothing vs.\ face binding.} General multi-condition models including LoomVideo-5B, MAGREF-14B, and VACE-14B generate roughly accurate person counts and reference clothing colors while scoring very low in \textbf{Face}, indicating that count and outfit alignment alone do not reflect facial identity preservation.
    \item \textbf{Identity preservation vs.\ action following.} Dedicated open-source reference-to-video models, namely UniVideo-13B, VINO-13B, and Phantom-14B, substantially improve \textbf{Face} and \textbf{IDCov}, while their \textbf{Action} scores remain modest. Clear gains in person-specific action following emerge mainly with larger backbone capacity, as seen in MiniMax-H3-33B and commercial models.
    \item \textbf{Complementary strengths among commercial models.} Despite close overall scores, commercial systems show distinct priorities: Seedance-2.5 leads in facial similarity and temporal consistency with lower image quality, MiniMax-H3 leads in action following and count accuracy with lower facial similarity, and Wan-3.0 excels in identity coverage and garment similarity.
\end{itemize}

\input{tables/k_common_metrics}

\subsection{Action Coverage and Execution}
\label{app:action-analysis}

Low action scores in multi-person generation can come from omitting assigned people in the video or from generating inaccurate motions when they appear. Tab.~\ref{tab:action-coverage-all} separates these two factors using the decomposition in Appendix~\ref{app:action-coverage-all}, where \textit{Coverage} measures how often assigned roles and interactions are visibly present, and \textit{Conditional} rates motion execution quality once observed.

This breakdown clarifies how action performance scales across models. Commercial models achieve both high role visibility and strong conditional execution. In contrast, open-source baselines up to 14B stay clustered at lower conditional ratings across both general and reference-specialized architectures, with a substantial jump appearing at 33B in MiniMax-H3-33B. Across backbones, GroupRef improves \textbf{Action} scores at all three scales primarily by raising \textit{Coverage}, bringing more assigned participants into the scene while preserving the motion quality of each base model.

\input{tables/action_coverage_all}

\subsection{Baseline Reference Conditioning}
\label{app:eval-protocol}
\label{app:reference-adaptation}

Evaluating group sizes up to $K=10$ exceeds the native reference capacity of many reference-to-video models, while commercial APIs impose server-side interface limits. Because existing generators ingest visual references through diverse architectural interfaces, Tab.~\ref{tab:ref-adaptation} groups all 20 evaluated models into six conditioning protocols. Open-weight baselines use four training-free inference strategies without weight updates or auxiliary modules:
(1) \textit{Native Variable} models directly ingest variable-length portrait lists of size $K$;
(2) \textit{Sequence Extrapolation} models extend visual conditioning tokens along their native reference sequence axis;
(3) \textit{Tensor Extension} models expand fixed-slot reference tensors along the slot axis under native attention routing; and
(4) \textit{Spatial Canvas} models tile the $K$ portraits onto their composite input grid via a deterministic layout.
\textit{GroupRef} variants retain each backbone's native pathway augmented with the shared learnable index anchors from Sec.~\ref{sec:method}, and commercial systems are queried through their official multi-reference APIs. Across all evaluated systems, each scene prompt is accompanied by a shared boundary clause specifying that only the $K$ referenced individuals appear without extra bystanders and with naturally visible faces, standardizing group-size and visibility expectations at inference.

\begin{table*}[!tp]
\centering
\caption{Multi-reference conditioning protocols across evaluated systems.}
\label{tab:ref-adaptation}
\small
\setlength{\tabcolsep}{4pt}
\begin{tabularx}{\textwidth}{@{}>{\raggedright\arraybackslash}p{0.22\textwidth}>{\raggedright\arraybackslash}p{0.34\textwidth}>{\raggedright\arraybackslash}X@{}}
\toprule
\textbf{Protocol} & \textbf{Evaluated Systems} & \textbf{Conditioning Mechanism} \\
\midrule
Native Variable & LoomVideo, Kaleido, VINO, UniVideo, VACE, RefAlign & Direct ingestion of variable-length portrait lists \\
Sequence Extrapolation & Phantom, MiniMax-H3-33B & Token-sequence continuation along native reference axis \\
Tensor Extension & OmniWeaving, BindWeave & Inference-time slot expansion under native attention routing \\
Spatial Canvas & MAGREF & Deterministic tiling onto native composite grid canvas \\
Anchor-Augmented & GroupRef Variants & Native backbone pathway augmented with learned anchors \\
Commercial Pipeline & Proprietary Systems & Official server-side multi-reference conditioning APIs \\
\bottomrule
\end{tabularx}
\end{table*}

\section{Evaluation Protocol and Validation}
\label{app:metrics}

Our evaluation pipeline follows three stages: (1) reference-free person tracking, (2) global one-to-one role matching, and (3) computation of nine metrics across Identity Preservation and Group Composition (IC), Visual Quality (VQ), and Instruction Following (IF), as summarized in Tab.~\ref{tab:metrics}. Role matching grounds \textbf{Face}, \textbf{IDCov}, \textbf{Garment}, and \textbf{Action}, whereas \textbf{Count}, \textbf{TempID}, \textbf{IQ}, and \textbf{Aes} do not depend on reference matching. Appendices~\ref{app:failure-axes} and~\ref{sec:human_eval} further present per-metric visual diagnostics and human agreement validation.

\begin{table*}[t]
\centering
\small
\caption{Summary of GroupRefBench evaluation metrics and weights.}
\label{tab:metrics}
\begin{tabular}{@{}llllr@{}}
\toprule
\textbf{Dimension} & \textbf{Metric} & \textbf{Evaluation Target} & \textbf{Video-Level Formulation} & \textbf{Weight} \\
\midrule
Identity \& Group Comp. (IC) & Face & Pooled entity face embeddings & Mean cosine over matched entities & 20\% \\
 & IDCov & Matched facial similarities & Supported identities / $K$ & 9\% \\
 & Count & Retained participant tracks & Inverse mean relative error $(1+\overline{E})^{-1}$ & 10\% \\
 & TempID & Long-gap intra-entity face pairs & Mean cosine over eligible tracks & 10\% \\
 & Garment & Masked entity clothing regions & Per-entity temporal median, then match mean & 5\% \\
\midrule
Visual Quality (VQ) & IQ & Uniformly sampled video frames & Mean MUSIQ prediction & 8\% \\
 & Aes & Uniformly sampled video frames & Mean LAION aesthetic score & 8\% \\
\midrule
Instruction Following (IF) & Action & Role-tagged keyframes & Equal weight per requested role & 20\% \\
 & Other & Role-tagged keyframes & Equal weight per checklist item & 10\% \\
\bottomrule
\end{tabular}
\end{table*}

\subsection{Person Tracking and Role Assignment}
\label{app:identity-correspondence}

For video $n$ with $T_n$ frames ($t \in \{0, \ldots, T_n-1\}$) and $K_n$ requested references, let $N_n$ denote the number of tracked participants.

We extract frame-level body boxes and keypoints via YOLO11x-pose~\citep{jocher2024yolo11} alongside person masks from SAM3.1~\citep{carion2025sam3}. Duplicate detections and competing face--body associations are resolved within each frame, and observations are linked across frames via motion cues, bidirectional mask continuity, and DINOv3 appearance features~\citep{simeoni2025dinov3}. Co-occurring bodies prevent track merging, overlapping regions split only under multi-body evidence, and post-occlusion re-entries are re-linked by appearance continuity instead of creating new tracks. Spatial extent, temporal persistence, and scene-layout cues filter out background pedestrians and static portraits without using $K_n$ or reference images; unmatched principal tracks remain in $N_n$.

From these retained participant tracks, we construct pooled facial descriptors across the full video sequence. Specifically, SCRFD~\citep{guo2022scrfd} detects faces and landmarks across all frames containing retained participants, linking detections to body masks via one-to-one spatial matching. Both reference portraits and video faces undergo identical alignment and encoding via a KP-RPE ViT-B backbone~\citep{kim2024kprpe} trained with AdaFace~\citep{kim2022adaface} on the WebFace12M subset of WebFace260M~\citep{zhu2021webface260m}, where the highest-confidence detection in reference $j$ defines its unit descriptor $r_j$. For a tracked entity $q$, valid facial observations require detector confidence $d_{qt}\geq 0.5$ and bounding-box short side $p_{qt}\geq 12$\,px, and forming a pooled entity descriptor $u_q$ requires at least three valid observations including at least one with $p_{qt}\geq 20$\,px. Given unit frame embeddings $f_{qt}$, we compute $u_q$ via quality-weighted pooling:
\begin{equation}
\omega_{qt}=d_{qt}\min(1,p_{qt}/80),\qquad
u_q=\operatorname{norm}\!\left(\sum_t\omega_{qt}f_{qt}\right),
\label{eq:group-face-pooling}
\end{equation}
where $\operatorname{norm}(x)=x/\|x\|_2$.

Finally, we establish candidate role correspondence by combining facial similarity $F_{qj}=u_q^\top r_j$ with temporal garment similarity $G_{qj}$ from Eq.~(\ref{eq:clothing-similarity}). Let $b^F_{qj},b^G_{qj}\in\{0,1\}$ indicate the availability of each modality. Over pairs $(q,j)$ with at least one available cue, we define the joint affinity
\begin{equation}
S_{qj}=\frac{0.25b^F_{qj}F_{qj}+0.75b^G_{qj}G_{qj}}
{0.25b^F_{qj}+0.75b^G_{qj}}.
\label{eq:joint-correspondence}
\end{equation}
In dynamic multi-person scenes, facial views naturally vary across pose angles, camera distances, and brief occlusions, whereas clothing regions maintain larger spatial extent throughout whole-body trajectories; weighting garment cues higher thus stabilizes global role binding while leaving downstream \textbf{Face} and \textbf{IDCov} scored strictly on facial similarity alone. Using $S_{qj}$, we solve a lexicographic Hungarian assignment~\citep{kuhn1955hungarian} that maximizes match cardinality prior to total affinity:
\begin{equation}
\mathcal M^*=\mathop{\arg\max}_{\mathcal M\;\mathrm{one\text{-}to\text{-}one}}
\left(|\mathcal M|,\sum_{(q,j)\in\mathcal M}S_{qj}\right)_{\mathrm{lex}}.
\label{eq:identity-matching}
\end{equation}
We denote the entity matched to reference $j$ as $q_j$, or $\emptyset$ if unmatched. This unthresholded one-to-one mapping establishes role correspondence while preserving $N_n$ for downstream evaluation.

\subsection{Identity, Group Composition, and Visual Quality}
\label{app:identity-metrics}
\label{app:quality-metrics}

Building upon the extracted tracks and role correspondences, we evaluate five complementary metrics for Identity Preservation and Group Composition alongside two frame-level Visual Quality metrics.

\noindent\textbf{Facial Fidelity and Coverage (Face \& IDCov).}
Let $\mathcal J_n^F$ denote the subset of references assigned to an entity $q_j\neq\emptyset$ with a measurable pooled face descriptor. Video-level facial fidelity and identity coverage are defined as
\begin{equation}
F_n=\frac{1}{|\mathcal J_n^F|}\sum_{j\in\mathcal J_n^F}F_{nq_jj},\qquad
C_n=\frac{1}{K_n}\sum_{j\in\mathcal J_n^F}\mathbf1[F_{nq_jj}\geq\tau].
\label{eq:identity-similarity-coverage}
\end{equation}
Here, \textbf{Face} ($F_n$) averages cosine similarities over measurable matches, whereas \textbf{IDCov} ($C_n$) retains all $K_n$ requested references in its denominator and applies a verification threshold $\tau$ to quantify identity coverage.

\noindent\textbf{Count Accuracy (Count).}
Given the per-video relative group-size error $E_n=|N_n-K_n|/K_n$, the dataset-level \textbf{Count} score across a video set $\mathcal V$ maps the mean relative error to $(0,1]$ via
\begin{equation}
\operatorname{Count}(\mathcal V)=\left(1+\frac{1}{|\mathcal V|}\sum_{n\in\mathcal V}E_n\right)^{-1},
\label{eq:count-score}
\end{equation}
yielding a smooth, bounded measure of group-size accuracy.

\noindent\textbf{Temporal Identity Consistency (TempID).}
To measure temporal facial stability independently of reference matching, we evaluate \textbf{TempID} across all tracked principal participants. For entity $q$, let $\mathcal V_{nq}$ collect frames with face confidence $\geq 0.5$ and short side $\geq 20$\,px. An entity qualifies for evaluation, denoted $q\in\mathcal Q_n^T$, when $|\mathcal V_{nq}|\geq 3$, its trajectory spans at least $0.4(T_n-1)$ frames, and it yields at least three long-gap frame pairs in
\begin{equation}
\mathcal P_{nq}=\{(t,t'):t,t'\in\mathcal V_{nq},\ t<t',\ t'-t\geq0.2(T_n-1)\}.
\label{eq:temporal-pairs}
\end{equation}
\textbf{TempID} averages pairwise cosine similarities equally across qualified entities:
\begin{equation}
U_n=\frac{1}{|\mathcal Q_n^T|}\sum_{q\in\mathcal Q_n^T}
\frac{1}{|\mathcal P_{nq}|}\sum_{(t,t')\in\mathcal P_{nq}}f_{nqt}^{\top}f_{nqt'},
\label{eq:temporal-coverage}
\end{equation}
directly complementing reference-grounded \textbf{Face} with internal trajectory consistency.

\noindent\textbf{Garment Similarity (Garment).}
To evaluate clothing preservation, we compare masked garment representations between references and matched entities. Each reference portrait is segmented via SAM3.1~\citep{carion2025sam3} using the prompt \texttt{person}, retaining the mask overlapping at least half of the reference face box. In generated frames, we isolate clothing regions on face-anchored person detections by subtracting the expanded face box from the tracked person mask. Each person crop is letterboxed to $448\times448$\,px and encoded by DINOv3 ViT-L/16~\citep{simeoni2025dinov3} without class or register tokens. Spatial patch embeddings $z_p$ are pooled post-encoding via garment-mask overlap weights $\alpha_p$, preserving contextual receptive fields while isolating clothing features:
\begin{equation}
\phi=\operatorname{norm}\!\left(\sum_p\alpha_pz_p\right).
\label{eq:clothing-feature}
\end{equation}
For each entity $q$, we construct an observation set $\mathcal T_{nq}$ of up to $16$ valid clothing crops by selecting the midpoint of each occupied shot and iteratively filling the largest temporal gaps. Over references $\mathcal J_n^G$ with measurable matched pairs, \textbf{Garment} computes the per-entity temporal median followed by the cross-reference mean:
\begin{equation}
\begin{gathered}
G_{nq_jj}=\operatorname{median}_{t\in\mathcal T_{nq_j}}
(\phi^{\mathrm{ref}}_{nj})^\top\phi^{\mathrm{gen}}_{nq_jt},\\
G_n=\frac{1}{|\mathcal J_n^G|}\sum_{j\in\mathcal J_n^G}G_{nq_jj}.
\end{gathered}
\label{eq:clothing-similarity}
\end{equation}

\noindent\textbf{Visual Quality (IQ \& Aes).}
Visual quality is evaluated independently of person tracking across uniformly sampled video frames. We extract $16$ frames at segment midpoints $\min(T_n-1,\operatorname{round}((j+0.5)T_n/16))$ for $j\in\{0,\ldots,15\}$ and resize their short side to $480$\,px. Image quality (\textbf{IQ}, $M_n$) averages KonIQ-10k MUSIQ scores~\citep{ke2021musiq,hosu2020koniq10k} via PyIQA~\citep{chen2022pyiqa}, while aesthetic quality (\textbf{Aes}, $A_n$) averages LAION aesthetic predictions~\citep{schuhmann2022aesthetic} over CLIP ViT-L/14 features~\citep{radford2021clip}.

\subsection{Person-Specific Instruction Following}
\label{app:instruction-metrics}
\label{app:action-coverage-all}

To evaluate instruction following, we score a prompt-specific checklist grounded in the preceding role-assignment stage (Appendix~\ref{app:evaluation-prompts}). Each checklist splits requirements into \textbf{Action} for individual actions and multi-person interactions and \textbf{Other} for scene context, objects, and layout, excluding person count and face identity. We sample $16$ chronological frames at $\operatorname{round}(i(T_n-1)/15)$ for $i\in\{0,\ldots,15\}$, resize them within $1024\times1024$\,px, and draw bounding boxes labeled with persistent role tags $R_j$ for \texttt{@Image j}, even during brief face occlusions.

GPT-5.5~\citep{openai2026gpt55} scores each checklist item $r_{ni}\in\{1,\ldots,5\}$ from these role-tagged frames when the involved people are visible, assigning low scores to incorrect actions. Checking consistency against Qwen3.8-Max~\citep{alibaba2026qwen38max} as a secondary judge shows high agreement at both the video level ($\rho = 0.866$ for \textbf{Action}, $0.831$ for \textbf{Other}) and in system rankings across the 20 models ($\rho = 0.976$ and $0.995$).

To aggregate checklist ratings into video-level scores, we apply tailored weighting schemes to the two categories. \textbf{Action} distributes weight equally across all $P_n$ requested roles carrying behavioral clauses, including omitted roles, and subdivides each role's share uniformly across its $L_{np}$ items, whereas \textbf{Other} weights all $L_n^{\mathrm{other}}$ items uniformly:
\begin{equation}
w^{\mathrm{act}}_{np\ell}=\frac{1}{P_nL_{np}},\qquad
w^{\mathrm{other}}_{n\ell}=\frac{1}{L_n^{\mathrm{other}}}.
\label{eq:requirement-weights}
\end{equation}
Let $\mathcal J_{na}$ denote the subset of visually verifiable items rated in judge evaluation $a$. The normalized evidence score $d_{na}$ and judgment coverage $c_{na}$ are computed as
\begin{equation}
d_{na}=\sum_{i\in\mathcal J_{na}}w_{ni}\frac{r_{nia}-1}{4},\qquad
c_{na}=\sum_{i\in\mathcal J_{na}}w_{ni}.
\label{eq:instruction-evidence}
\end{equation}

\begin{figure*}[t]
\centering
\includegraphics[width=\textwidth]{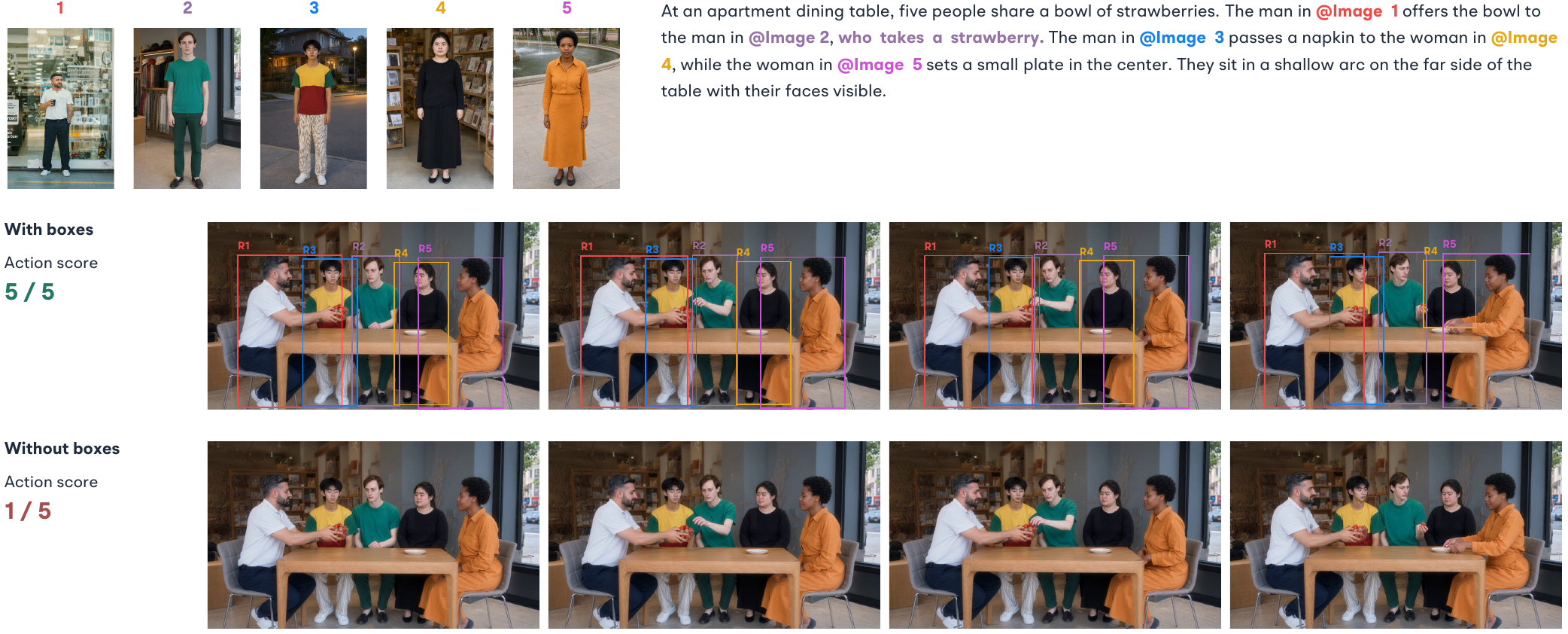}
\caption{Action scoring on the same video with and without role-tagged boxes $R_j$.}
\label{fig:vlm-bbox-case}
\end{figure*}

\noindent\textbf{Effect of Role-Tagged Boxes.}
Overlaying persistent role tags directly grounds VLM action judgments in spatial trajectories, as shown in Fig.~\ref{fig:vlm-bbox-case}. With role-tagged boxes, the judge tracks \texttt{@Image 2} taking a strawberry and assigns a $5/5$ rating; without them, \texttt{@Image 2} lacks spatial grounding and is misattributed to another participant whose hands remain away from the bowl, yielding a $1/5$ rating. Projecting persistent role tags onto frames thus helps prevent actor misattribution when evaluating person-specific behaviors.

\noindent\textbf{Action Score Decomposition.}
To distinguish missing role appearances from inaccurate motion synthesis, we decompose \textbf{Action} into \textbf{Judgment Coverage}, measuring whether requested behaviors and interaction partners are visually observable under assigned roles, and \textbf{Conditional Rating}, measuring execution fidelity on the $1$--$5$ scale once observed, as reported in Tab.~\ref{tab:action-coverage-all}. Averaging $c_{na}$ and $d_{na}$ across judge evaluations yields per-video $\bar c_n$ and $\bar d_n$. Over $M$ valid videos, dataset-level coverage and conditional rating are defined as
\begin{equation}
C=\frac{1}{M}\sum_n\bar c_n,\qquad
R_{\mathrm{cond}}=1+4\frac{\sum_n\bar d_n}{\sum_n\bar c_n},
\end{equation}
admitting the exact multiplicative decomposition
\begin{equation}
\mathrm{Action}=20C\frac{R_{\mathrm{cond}}-1}{4},
\end{equation}
where $R_{\mathrm{cond}}$ represents a coverage-weighted aggregate across the evaluated videos.

\subsection{Score Aggregation and Evaluation Scopes}
\label{app:score-aggregation}
\label{app:video-availability}

All models are evaluated under two observation scopes: all returned videos per model in Tab.~\ref{tab:grouprefbench-results}, and the shared $N=229$ video intersection in Appendix~\ref{app:k-coverage} to control for API filtering and 9-image input limits. Within each scope, video-level completeness and quality metrics (\textbf{Count}, \textbf{IDCov}, \textbf{IQ}, \textbf{Aes}) are averaged across all evaluated videos, whereas conditional similarity metrics (\textbf{Face}, \textbf{Garment}, \textbf{TempID}) are averaged over valid detections, and \textbf{Action} combines visibility coverage with conditional action ratings.

Denoting the resulting model-level marginal means across the $N$ evaluated videos as $\overline F, \overline C, \overline E, \overline U, \overline G, \overline M, \overline A, D^{\mathrm{act}}$, and $D^{\mathrm{other}}$, the overall benchmark score (\textbf{Total}) combines all nine components via fixed weights prior to rounding:
\begin{equation}
\begin{split}
\operatorname{Total}={}&20\overline F+9\overline C+
\frac{10}{1+\overline E}+10\overline U+5\overline G\\
&+0.08\overline M+0.8\overline A+
20D^{\mathrm{act}}+10D^{\mathrm{other}},
\end{split}
\label{eq:overall-evidence}
\end{equation}
where $\overline M$ (\textbf{IQ}) and $\overline A$ (\textbf{Aes}) are normalized by $100$ and $10$ before applying their $8\%$ weights, and cosine similarity means $\overline F, \overline G, \overline U$ preserve their signed values.

\subsection{Per-Metric Case Studies}
\label{app:failure-axes}

\begin{figure*}[!tp]
\centering
\includegraphics[width=\textwidth]{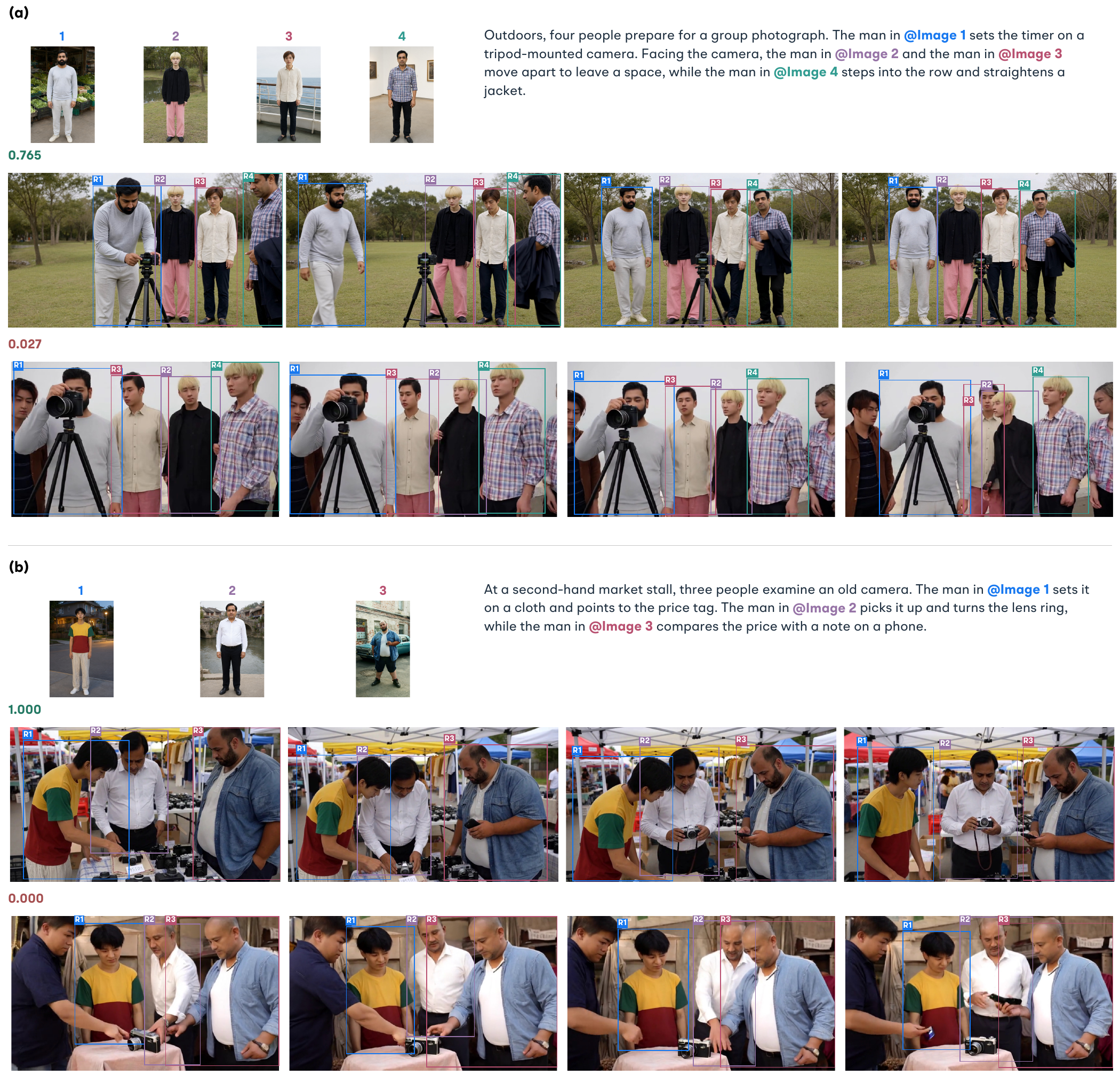}
\caption{Qualitative examples for (a) \textbf{Face} and (b) \textbf{IDCov}. $R_j$ denotes matched reference roles.}
\label{fig:selected-identity}
\end{figure*}

\begin{figure*}[!tp]
\centering
\includegraphics[width=\textwidth]{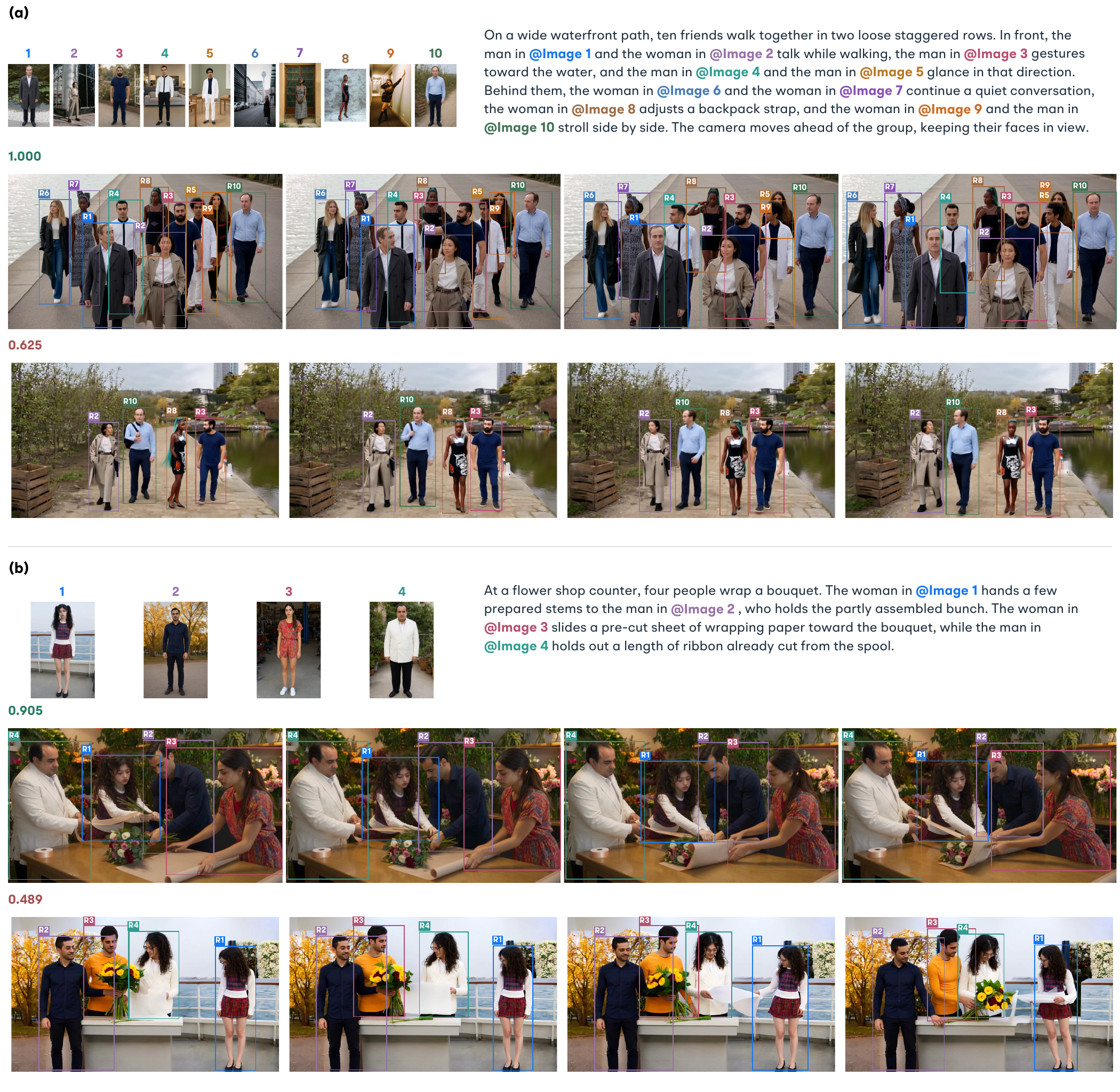}
\caption{Qualitative examples for (a) \textbf{Count} and (b) \textbf{TempID}.}
\label{fig:selected-count-temporal}
\end{figure*}

\begin{figure*}[!tp]
\centering
\includegraphics[width=\textwidth]{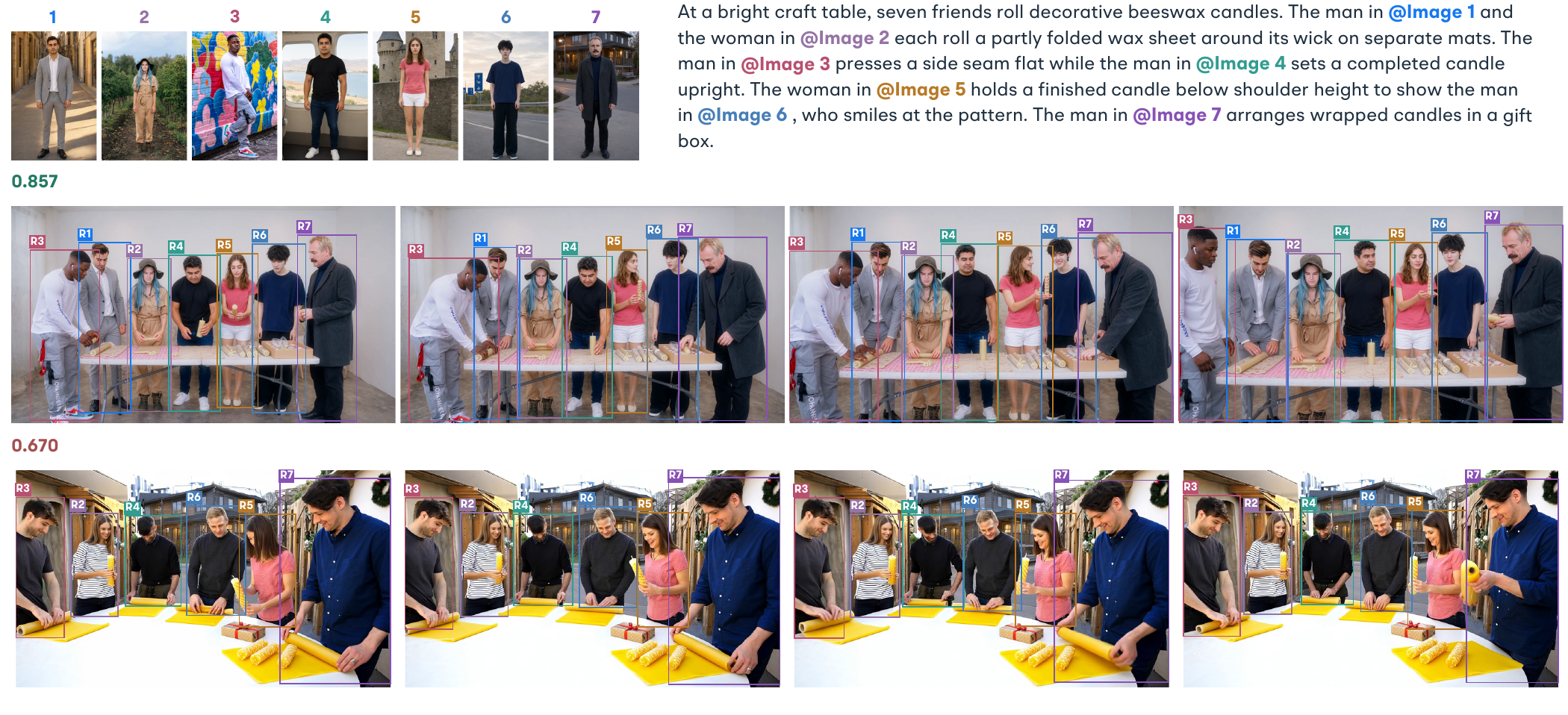}
\caption{Qualitative example for \textbf{Garment} on a seven-person scene.}
\label{fig:selected-garment}
\end{figure*}

To illustrate how our five Identity Preservation and Group Composition metrics capture different failure modes, Figs.~\ref{fig:selected-identity}--\ref{fig:selected-garment} pair high-scoring (top) and low-scoring (bottom) generations from identical prompts and references. Four frames are shown per video, while scores are computed over the full sequence using matched role tags $R_j$.

\noindent\textbf{Face and IDCov.}
Evaluating identity preservation requires measuring both continuous facial similarity and thresholded identity coverage, as compared in Fig.~\ref{fig:selected-identity} for \textbf{Face} and \textbf{IDCov}.
In Fig.~\ref{fig:selected-identity}(a), both videos generate all four people in matching outfits and scene layouts, yet the bottom video synthesizes plausible faces of the wrong identities.
In the three-person market scene of Fig.~\ref{fig:selected-identity}(b), the bottom video adds an extra unreferenced person on the left and matches $R_1$--$R_3$ by shirt color, while none of the matched faces passes the verification threshold $\tau$.
Reporting \textbf{Face} and \textbf{IDCov} together thus separates facial identity preservation from clothing colors and person counts.

\noindent\textbf{Count and TempID.}
Multi-person generation also requires rendering the right number of people and keeping each person's face stable over time, as shown in Fig.~\ref{fig:selected-count-temporal} for \textbf{Count} and \textbf{TempID}.
In the ten-person scene of Fig.~\ref{fig:selected-count-temporal}(a), the top video renders all ten people, whereas the bottom video generates only four walkers despite high visual quality.
In the four-person bouquet scene of Fig.~\ref{fig:selected-count-temporal}(b), all four tracks remain clearly separated, yet faces in the bottom video drift noticeably across frames.
Measuring long-gap similarity within each track isolates temporal facial drift from reference mismatch and missing subjects.

\noindent\textbf{Garment Similarity.}
Clothing consistency provides an additional signal for person-level fidelity, which we illustrate with \textbf{Garment} on a seven-person scene in Fig.~\ref{fig:selected-garment}.
The top video preserves the distinctive outfits of all seven people, whereas the bottom video omits $R_1$ and replaces characteristic outfits, such as $R_2$'s yellow overalls and $R_3$'s white hoodie, with plain shirts.
Overall, Figs.~\ref{fig:selected-identity}--\ref{fig:selected-garment} show how these five metrics diagnose distinct errors that visual quality metrics alone miss.

\subsection{Human Validation}
\label{sec:human_eval}

\begin{figure*}[!tp]
\centering
\includegraphics[width=\textwidth]{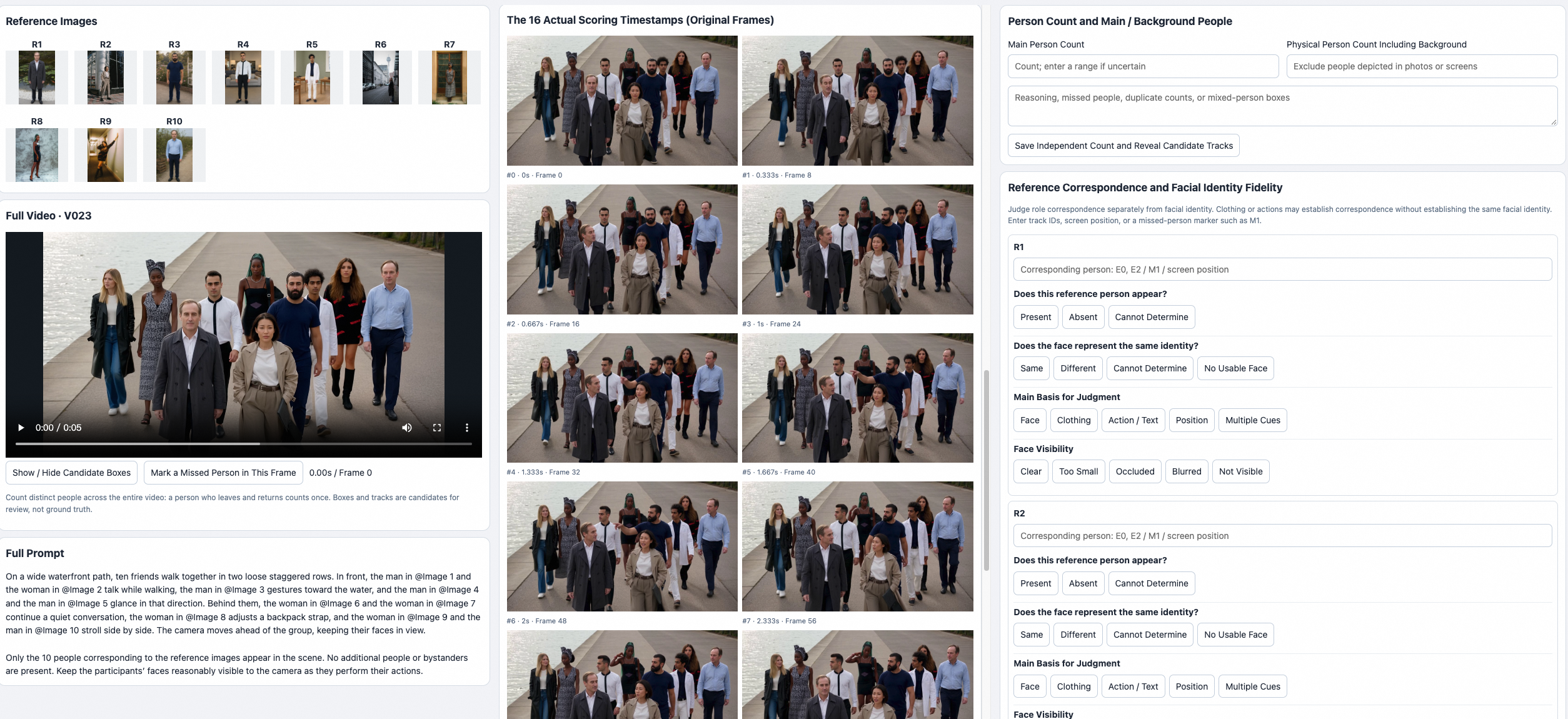}
\caption{Human annotation interface for participant count and identity correspondence.}
\label{fig:human_eval}
\end{figure*}

\begin{figure*}[!tp]
\centering
\includegraphics[width=\textwidth]{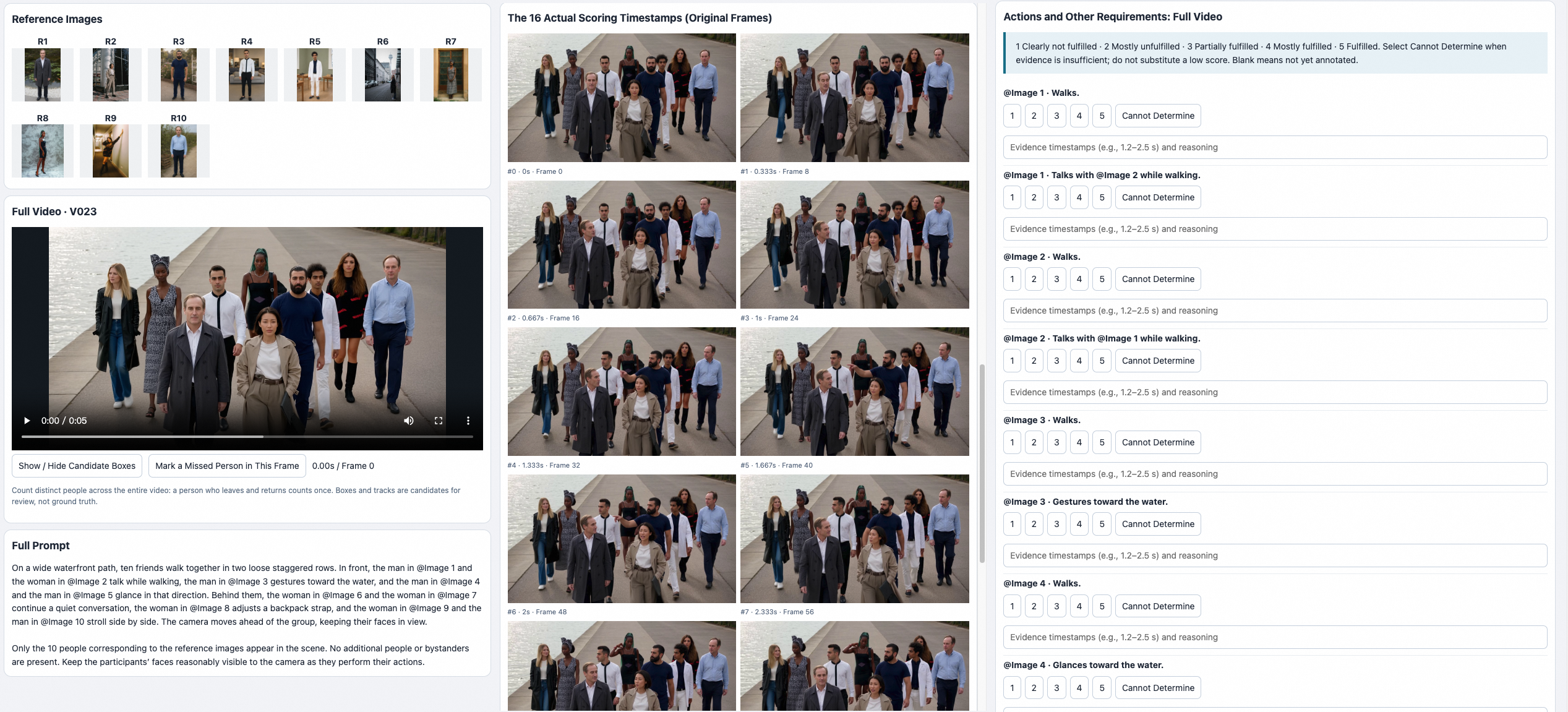}
\caption{Human annotation interface for \textit{Action} and \textit{Other} checklists.}
\label{fig:human_eval_actions}
\end{figure*}

To check how well our automated evaluation aligns with human judgment, annotators reviewed 480 generated videos on GroupRefBench using the interfaces in Figs.~\ref{fig:human_eval} and~\ref{fig:human_eval_actions}.

\begin{table*}[!tp]
\centering
\caption{Human--pipeline agreement on GroupRefBench.}
\label{tab:human_agreement}
\small
\setlength{\tabcolsep}{12pt}
\begin{tabular}{@{}lllr@{}}
\toprule
\textbf{Evaluation dimension} & \textbf{Granularity} & \textbf{Metric} & \textbf{Agreement} \\
\midrule
Person Count & Video & MAE $\downarrow$ / Exact match $\uparrow$ & 0.06 / 97.5\% \\
Role Assignment & Reference slot & Accuracy $\uparrow$ & 94.2\% \\
Face Identity & Reference slot & ROC-AUC $\uparrow$ & 0.828 \\
Action IF & Checklist item & Spearman's $\rho$ $\uparrow$ & 0.776 \\
Other IF & Checklist item & Spearman's $\rho$ $\uparrow$ & 0.848 \\
\bottomrule
\end{tabular}
\end{table*}

As summarized in Tab.~\ref{tab:human_agreement}, automated outputs align well with human annotations across all five dimensions:
\begin{itemize}
    \item \textbf{Person Count and Role Assignment.} Predicted participant counts closely match human counts of main participants, excluding background passersby, and bipartite role matching agrees well with human-verified role assignments.
    \item \textbf{Face Identity.} Pooled face similarity scores effectively distinguish human-verified identity matches from mismatches across reference slots.
    \item \textbf{IF Checklist Scores.} Role-tagged VLM scores show consistent rank correlation with human ratings on both the \textit{Action} and \textit{Other} checklists.
\end{itemize}

\section{Benchmark Construction and Comparison}

This section details the portrait and prompt curation process of GroupRefBench (Appendix~\ref{app:bench-construction}) and compares its input conventions and evaluation protocol against existing reference-to-video benchmarks (Appendix~\ref{app:benchmark-comparison}).

\subsection{GroupRefBench Construction}
\label{app:bench-construction}

Supplementing Sec.~\ref{sec:bench-construction}, each GroupRefBench sample pairs $K \in \{3, \ldots, 10\}$ distinct reference portraits from Pexels~\citep{pexels} and Qwen-Image-3.0-Pro~\citep{qwen2026qwenimage3} with an appearance-free prompt expanded by Qwen3.8-Max~\citep{alibaba2026qwen38max} from human-designed scene seeds. Every prompt assigns an action or interaction to each of the $K$ indices without describing faces, hairstyles, or clothing, followed by automated tag checks and manual review.

As shown in Fig.~\ref{fig:dataset-overview}(a, b), prompt length grows naturally with group size $K$ as more individual roles are specified, while the interaction structure shifts from independent actions in smaller groups toward joint actions and social interactions in larger groups.

\subsection{Comparison with Existing Benchmarks}
\label{app:benchmark-comparison}
\label{app:correspondence-comparison}

To situate GroupRefBench among existing evaluation suites, Tab.~\ref{tab:benchmark-comparison} summarizes the input conventions and evaluation protocols of representative reference-to-video benchmarks. In most existing datasets, nominal reference counts reflect input interface slots instead of distinct human identities. Multiple slots frequently supply multi-view crops of a single subject alongside an optional background image, as in MSRVTT-Personalization~\citep{chen2025videoalchemist} and IntelligentVBench~\citep{pan2026omniweaving}, whereas composite images in MultiRef-Compass~\citep{zhang2026multirefcompass} and OmniVBench~\citep{omnivbench2026} pack multiple visual entities or distractors into fewer slots, and FashionVideoBench~\citep{wu2026loomvideo} leaves its reference capacity unspecified. In terms of prompt addressing, index tags often coexist with descriptive cues to disambiguate targets within composite inputs; accordingly, the \emph{Appearance} column in Tab.~\ref{tab:benchmark-comparison} indicates whether descriptive cues are permitted, without requiring exhaustive facial or wardrobe specifications.

\noindent\textbf{Identity Evaluation.}
Existing benchmarks evaluate human identity through three paradigms: single-subject verification, framewise multi-person matching, and holistic VLM rubrics. Single-subject protocols in MSRVTT-Personalization~\citep{chen2025videoalchemist} and OpenS2V-Eval~\citep{yuan2025opens2vnexus} assume at most one target human per video, delegating broader appearance assessment to general visual similarity metrics. Specifically, MSRVTT-Personalization pairs YOLOv9-C with ArcFace R100 to score the first detected face per frame while assigning zero to frames without detections, whereas OpenS2V-Eval pairs InsightFace with CurricularFace IR-101 to average strictly positive similarities over the largest detected face per frame. Extending embedding comparisons to multiple humans, MultiRef-Compass~\citep{zhang2026multirefcompass} extracts framewise ElasticFace R100 features $F$ and SigLIP body features $B$, ranking pairs by $0.4F+0.6B$ for greedy per-frame selection and scoring matched pairs via $0.6F+0.4B$, with unmatched references receiving a default score of $0.5$.

Conversely, UniVBench~\citep{wei2026univbench}, IntelligentVBench~\citep{pan2026omniweaving}, FashionVideoBench~\citep{wu2026loomvideo}, and OmniVBench~\citep{omnivbench2026} evaluate reference consistency through holistic VLM rubrics without face embeddings or explicit entity assignment. These protocols prompt multimodal judges to jointly assess facial likeness, wardrobe, subject count, and scene elements. Even with targeted checks for identity or clothing swaps in IntelligentVBench and factorized verification queries in OmniVBench, all four benchmarks rely on implicit visual reasoning instead of algorithmic person-to-reference correspondence.

\begin{table*}[t]
\centering
\small
\caption{Comparison of multi-person correspondence protocols.}
\label{tab:correspondence-comparison}
\setlength{\tabcolsep}{5pt}
\renewcommand{\arraystretch}{1.12}
\begin{tabularx}{\textwidth}{@{}>{\raggedright\arraybackslash}p{0.22\textwidth}>{\raggedright\arraybackslash}X>{\raggedright\arraybackslash}X@{}}
\toprule
\textbf{Design Aspect} & \textbf{MultiRef-Compass~\citep{zhang2026multirefcompass}} & \textbf{GroupRefBench} \\
\midrule
Matching unit & Per-frame detection candidate & Cross-frame tracked entity \\
Face-to-body binding & Independent face enclosure per body box & Exclusive one-to-one spatial association \\
Temporal strategy & Independent per-frame assignment & Reference-free tracking then global matching \\
Matching cues & Framewise ElasticFace and SigLIP features & Pooled face and temporal garment descriptors \\
Assignment rule & Greedy selection per frame & Lexicographic Hungarian optimization \\
Face scoring & Mean over retained framewise matches & Cosine similarity on pooled entity descriptors \\
\bottomrule
\end{tabularx}
\end{table*}

\noindent\textbf{Multi-Person Correspondence.}
Multi-person identity evaluation depends on how video detections are associated with input references; Tab.~\ref{tab:correspondence-comparison} compares the design choices of the framewise matching protocol in MultiRef-Compass~\citep{zhang2026multirefcompass} and the video-level protocol in GroupRefBench. Under framewise greedy selection, each frame is matched independently without trajectory continuity, so pose changes, overlapping detections, or shot transitions can occasionally lead to duplicate or shifted assignments in crowded scenes. Tracking trajectories before solving a single video-level assignment binds each reference to one trajectory across frames, while pairing unthresholded match similarity in \textbf{Face} with full-group normalization in \textbf{IDCov} separates conditional facial similarity from overall identity coverage.

\begin{figure*}[t]
\centering
\includegraphics[width=\textwidth]{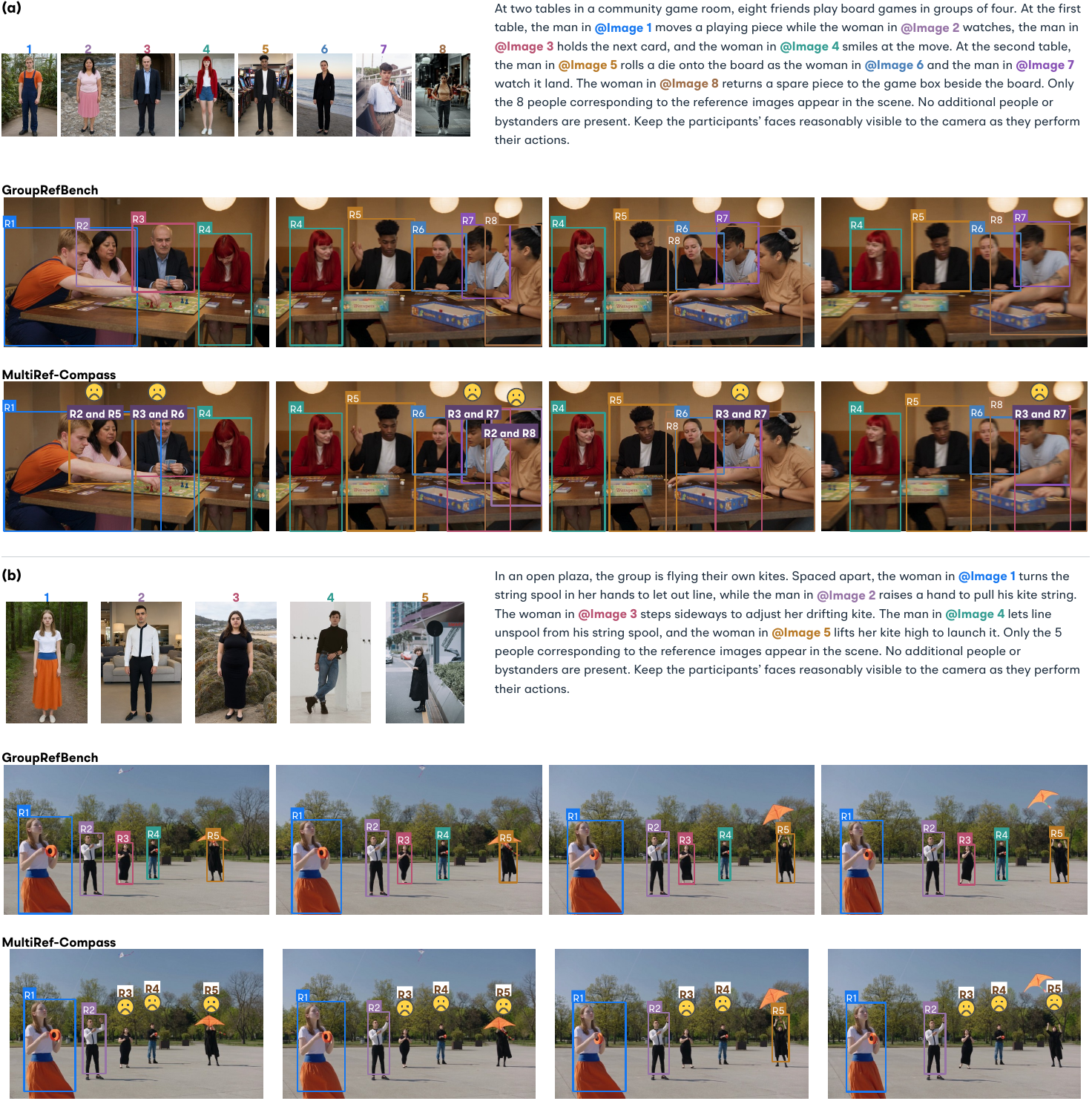}
\caption{Illustrative comparison of framewise and video-level reference correspondence on (a) a shot transition and (b) distant participants.}
\label{fig:correspondence-cases}
\end{figure*}

Fig.~\ref{fig:correspondence-cases} illustrates observed assignment behaviors on two example Seedance videos when comparing framewise matching against our video-level protocol. In the two-table scene of Fig.~\ref{fig:correspondence-cases}(a), framewise assignment matches overlapping regions to multiple references and reassigns $R_2$ and $R_3$ after the shot cut, whereas video-level tracking leaves off-screen roles unboxed during those frames. In Fig.~\ref{fig:correspondence-cases}(b), several distant participants remain unmatched on individual frames under framewise selection, whereas video-level matching maintains role assignments across the sequence by aggregating multi-frame cues.

\noindent\textbf{Action Evaluation.}
Beyond low-level motion smoothness metrics, existing benchmarks evaluate requested behaviors through static caption alignment, holistic video ratings, or VLM checklists, as summarized in Tab.~\ref{tab:benchmark-comparison}. At the coarsest granularity, MSRVTT-Personalization~\citep{chen2025videoalchemist} and OpenS2V-Eval~\citep{yuan2025opens2vnexus} average frame-to-prompt feature similarities via CLIP ViT-L/14 and GME-Qwen2-VL-7B-Instruct, measuring overall video--text alignment without isolating individual actions or verifying actor identity. Using multimodal judges, UniVBench~\citep{wei2026univbench}, IntelligentVBench~\citep{pan2026omniweaving}, and FashionVideoBench~\citep{wu2026loomvideo} summarize behavioral adherence in a single video-level score via Seed-1.6 or Gemini 2.5 Pro, combining action execution with motion dynamics, camera movement, and scene composition without reporting per-subject scores.

For fine-grained actor attribution, MultiRef-Compass~\citep{zhang2026multirefcompass} and OmniVBench~\citep{omnivbench2026} pair Gemini 3.1 Pro with structured checklists. MultiRef-Compass separates actor-agnostic action occurrence from categorical binding without forwarding its feature-level entity matches to the VLM, whereas OmniVBench evaluates reference-specific checklist items directly from visual inputs and metadata. Both protocols therefore ask the VLM to infer identity correspondence and behavioral execution jointly. GroupRefBench instead projects persistent role tags from global tracking and matching onto keyframes prior to VLM evaluation, as detailed in Appendices~\ref{app:identity-correspondence} and~\ref{app:instruction-metrics}. Grounded in the preceding role-assignment stage, \textbf{Action} evaluates behavioral execution on the matched trajectories without directly conditioning on reference images or facial similarity scores.

\section{GroupRef100K Construction}
\label{app:data}

GroupRef100K curates $100{,}000$ multi-person training samples from $500{,}000$ source clips through the pipeline summarized in Tab.~\ref{tab:data-pipeline}, corresponding to the three stages in Sec.~\ref{sec:dataset}: Reference Curation (Appendix~\ref{app:data-curation}), Action Grounding (Appendix~\ref{app:data-caption}), and Reference Diversification (Appendix~\ref{app:data-diversification}). All evaluation samples in GroupRefBench and OpenS2V-Eval are constructed from independent image pools and prompts disjoint from this training collection. As shown in Fig.~\ref{fig:dataset-overview}(g), multi-person clips ($K \ge 3$) contribute most of the references, with augmented portraits accounting for roughly one-third of references in larger groups.

\begin{table*}[t]
\centering
\small
\caption{Data curation pipeline for GroupRef100K.}
\label{tab:data-pipeline}
\setlength{\tabcolsep}{5pt}
\renewcommand{\arraystretch}{1.12}
\begin{tabularx}{\textwidth}{@{}>{\raggedright\arraybackslash}p{0.20\textwidth}>{\raggedright\arraybackslash}p{0.36\textwidth}>{\raggedright\arraybackslash}X@{}}
\toprule
\textbf{Pipeline Stage} & \textbf{Core Models and Signals} & \textbf{Key Mechanism and Verification} \\
\midrule
Person tracking & Grounding DINO~\citep{liu2023grounding}, face detector, CLIP~\citep{radford2021clip} & $4$\,fps sampling, one-to-one spatial association, Hungarian linking~\citep{kuhn1955hungarian} \\
Reference selection & Frontality, sharpness, completeness, and occlusion cues & Per-track view ranking and distinctness filtering \\
Action grounding & Qwen3.8-Max~\citep{alibaba2026qwen38max} and tag parser & Caption-to-video alignment with \texttt{@Image k} completeness checks \\
Diversification & Visual similarity models, image editor, Qwen3.8-Max & Cross-shot view pairing and background editing \\
\bottomrule
\end{tabularx}
\end{table*}

\subsection{Reference Curation}
\label{app:data-curation}

Frames sampled at $4$\,fps are processed to detect participants and link observations across frames using visual appearance and spatial continuity. From each participant's candidate pool, we rank views by sharpness, frontality, body completeness, and minimal occlusion to select one clear reference image per individual.

\subsection{Action Grounding}
\label{app:data-caption}

Selected references are aligned with appearance-free instructions by matching reference crops to person mentions in the source caption while supplementing unmentioned visible participants from sampled frames. Qwen3.8-Max~\citep{alibaba2026qwen38max} rewrites the caption into a coherent instruction that replaces appearance descriptions with \texttt{@Image k} tags while preserving actions, interactions, spatial layout, and temporal order under automated tag checks.

\subsection{Reference Diversification}
\label{app:data-diversification}

To mitigate copy-paste artifacts while keeping target videos and \texttt{@Image k} prompts fixed, reference crops are diversified via cross-shot pairing or background editing. Augmented candidates replace the original crop only when they pass visual consistency and VLM quality checks.

\section{GroupRef Training and Analyses}
\label{app:model-analyses}

This section provides the training and OpenS2V-Eval setup for GroupRef (Appendix~\ref{app:training}), quantitative and visual analyses of reference attention discrimination (Appendix~\ref{app:attention}), and qualitative examples of input permutation consistency (Appendix~\ref{app:permutation}).

\subsection{Training and OpenS2V-Eval Setup}
\label{app:training}
\label{app:opens2v-eval}

We instantiate GroupRef across three backbone scales and evaluate out-of-distribution generalization on OpenS2V-Eval~\citep{yuan2025opens2vnexus}. Specifically, we initialize GroupRef-1.3B and GroupRef-14B from RefAlign-1.3B and RefAlign-14B~\citep{wang2026refalign} and GroupRef-33B from MiniMax-H3-33B~\citep{minimax2026h3}, keeping pretrained backbone weights frozen while training with LoRA~\citep{hu2022lora} and CMIA anchors on GroupRef100K.

On OpenS2V-Eval~\citep{yuan2025opens2vnexus}, we evaluate GroupRef-1.3B and GroupRef-14B across all 180 test cases conditioned on one to three reference images per sample. All results in Tab.~\ref{tab:opens2v-results} follow the official evaluation protocol, including metric definitions, score normalization, and weighted aggregation into the overall \textbf{Total} score. The benchmark evaluates four complementary dimensions:
(a) \emph{subject and facial fidelity}, measured by \textbf{NexusScore} for reference-to-region subject consistency and \textbf{FaceSim} for CurricularFace identity similarity;
(b) \emph{visual integration and aesthetics}, measured by \textbf{NaturalScore}, which penalizes copy-paste artifacts, and \textbf{AestheticScore};
(c) \emph{prompt alignment}, measured by \textbf{GmeScore}; and
(d) \emph{motion dynamics}, measured by \textbf{MotionSmoothness} and \textbf{MotionAmplitude}.
We compare GroupRef against proprietary systems including Kling-1.6~\citep{kling2025official}, Vidu-2.0~\citep{bao2024vidu}, and Pika-2.1~\citep{pika2025official}, as well as research methods including Saber~\citep{zhou2025saber}, VACE~\citep{jiang2025vace}, BindWeave~\citep{li2025bindweave}, VINO~\citep{chen2026vino}, Phantom~\citep{liu2025phantom}, MAGREF~\citep{deng2026magref}, SkyReels-A2~\citep{fei2025skyreelsa2}, and our base models RefAlign-1.3B and RefAlign-14B~\citep{wang2026refalign}, using published scores evaluated under the same official protocol.

\subsection{Reference Attention Analysis}
\label{app:attention}

\begin{figure*}[t]
\centering
\includegraphics[width=\linewidth]{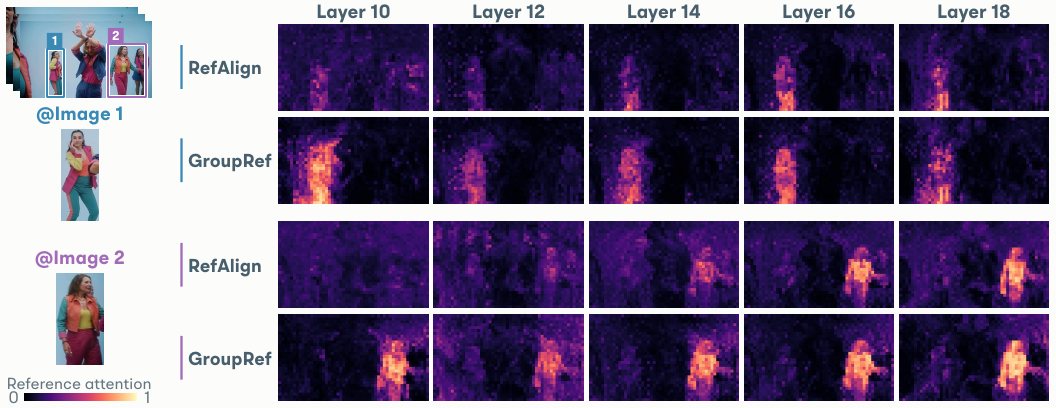}
\caption{Reference attention maps across DiT blocks.}
\label{fig:attnfig}
\end{figure*}

To complement the ablations in Sec.~\ref{sec:ablation}, we analyze reference attention discrimination across RefAlign-1.3B, GroupRef-1.3B w/o CMIA, and GroupRef-1.3B on $500$ test videos balanced across group sizes. Motivated by prior evidence that subject-aligned attention under low noise supports identity preservation~\citep{yu2026harnessing}, we inspect zero-indexed DiT blocks $10$--$18$ at a fixed low-noise timestep, extracting queries and keys after linear projection, normalization, and rotary positional encoding. Representative visualizations are shown in Fig.~\ref{fig:attnfig}.

Within each inspected block, let $q_{hu}$ denote the query at video-token position $u$ in head $h$, and $k_{hv}$ the key at token position $v$. The attention mass routed from video token $u$ to reference image $j$ is
\begin{equation}
a_{uj}
=
\frac{1}{H}
\sum_{h=1}^{H}
\sum_{v\in\mathcal{R}_j}
\frac{
    \exp(q_{hu}^{\top}k_{hv}/\sqrt{d_h})
}{
    \sum_{w\in\mathcal{S}}
    \exp(q_{hu}^{\top}k_{hw}/\sqrt{d_h})
},
\label{eq:reference-attention-mass}
\end{equation}
where $H$ is the number of heads, $d_h$ is the per-head dimension, $\mathcal{R}_j$ indexes the tokens of reference $j$, and $\mathcal{S}$ denotes the concatenated sequence of video and reference tokens. To quantify subject-level discrimination, we project annotated person masks onto the video-token grid and extract a nonempty foreground query set $\Omega_i$ for each subject $i$ from low-overlap regions, weighted by fixed token-level mask occupancy $w_{iu} > 0$. Given $K$ references, the token-level correct-reference margin $g_i(u)$ and its occupancy-weighted subject-level average $G_i$ are
\begin{equation}
\begin{aligned}
g_i(u)
&=
a_{ui}
-
\frac{1}{K-1}
\sum_{j\ne i}a_{uj},
\\
G_i
&=
\frac{
    \sum_{u\in\Omega_i}w_{iu}g_i(u)
}{
    \sum_{u\in\Omega_i}w_{iu}
}.
\end{aligned}
\label{eq:reference-attention-margin}
\end{equation}
For each video, $G_i$ is averaged uniformly across valid subjects and the nine inspected blocks. Tab.~\ref{tab:attention-three-models} reports the mean margin across all $500$ test videos, scaled by $100$ into percentage points, together with $95\%$ paired bootstrap confidence intervals.

\begin{table*}[t]
\centering
\small
\caption{Reference attention margins in percentage points across $500$ test videos, with $95\%$ bootstrap confidence intervals in superscripts.}
\label{tab:attention-three-models}
\setlength{\tabcolsep}{12pt}
\renewcommand{\arraystretch}{1.15}
\begin{tabular}{@{}lc@{}}
\toprule
\textbf{Model / Pairwise Comparison} & \textbf{Correct $-$ Mean Wrong $\uparrow$} \\
\midrule
RefAlign-1.3B & $1.4290$ \\
GroupRef-1.3B w/o CMIA & $4.2601$ \\
GroupRef-1.3B & $\mathbf{4.4023}$ \\
\midrule
GroupRef-1.3B w/o CMIA $-$ RefAlign-1.3B & $+2.8311^{[2.7053,\,2.9611]}$ \\
GroupRef-1.3B $-$ RefAlign-1.3B & $+2.9733^{[2.8463,\,3.1052]}$ \\
GroupRef-1.3B $-$ GroupRef-1.3B w/o CMIA & $+0.1422^{[0.1364,\,0.1482]}$ \\
\bottomrule
\end{tabular}
\end{table*}

Because $a_{uj}$ in Eq.~(\ref{eq:reference-attention-mass}) is normalized over the full token sequence $\mathcal{S}$ and averaged across all heads, reference attention values span a compact numerical range. As shown in Fig.~\ref{fig:attnfig} and Tab.~\ref{tab:attention-three-models}, training on GroupRef100K substantially increases correct-reference attention over RefAlign-1.3B, and incorporating CMIA further sharpens subject-specific attention routing with a positive confidence interval.

\subsection{Input Permutation Consistency}
\label{app:permutation}

\begin{figure*}[!t]
\centering
\includegraphics[width=\textwidth]{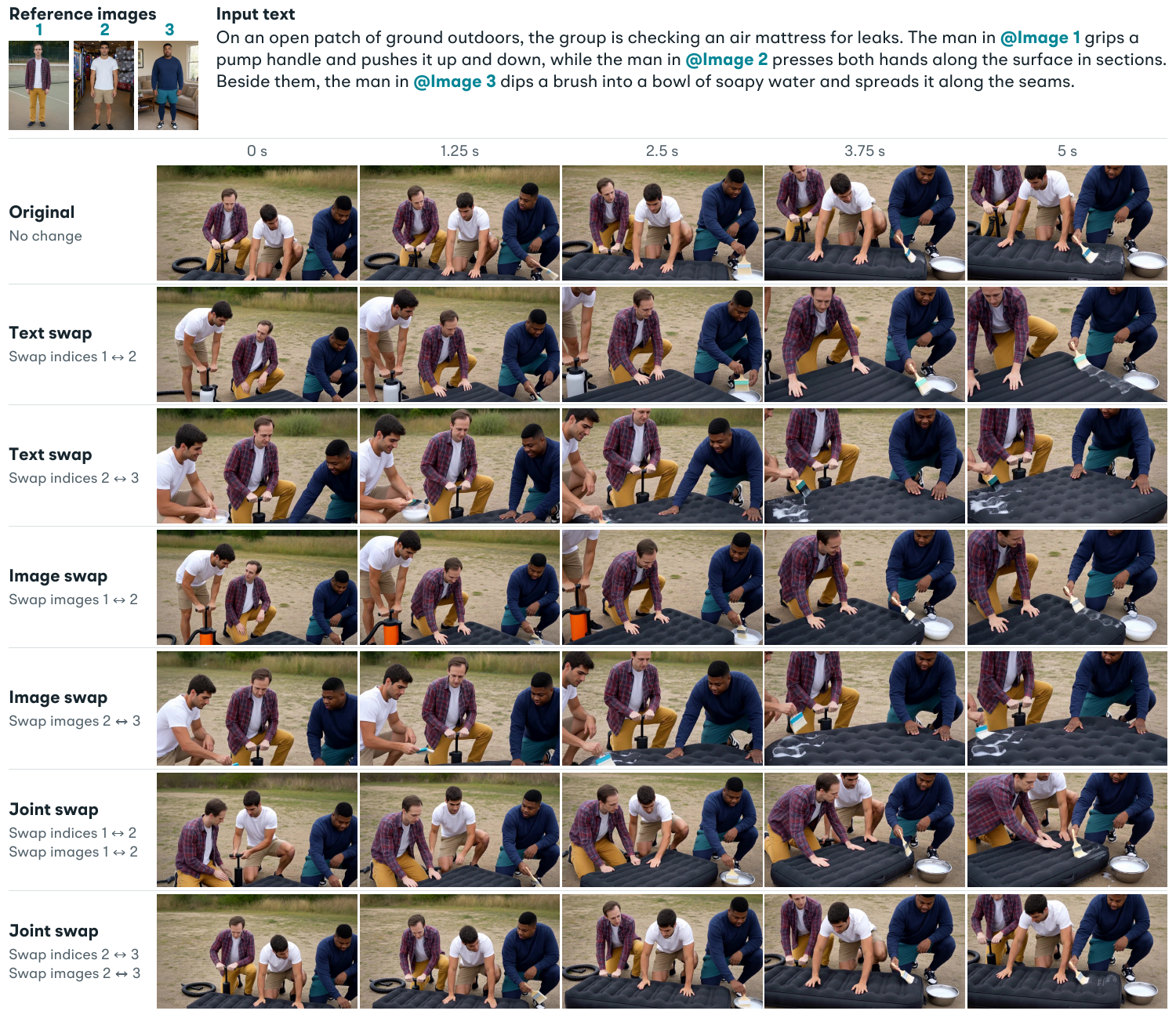}
\caption{Counterfactual permutations of prompt indices, reference images, and both modalities jointly.}
\label{fig:permutation}
\end{figure*}

To examine whether the model binds people by explicit indices or relies on input order, we swap pairs of indices or images ($1\leftrightarrow2$ and $2\leftrightarrow3$) on an example scene under three settings: swapping text indices only, swapping reference images only, and swapping both together. As shown in Fig.~\ref{fig:permutation}, swapping a single modality reassigns the corresponding people to the newly matched actions, whereas swapping both together keeps each person's action unchanged, providing a qualitative illustration of index-based binding under different reference permutations.

\clearpage
\onecolumn
\raggedbottom
\section{Evaluation Prompts}
\label{app:evaluation-prompts}

The boxes below summarize the checklist-extraction and joint-scoring system prompts used in our evaluation pipeline.
Each checklist is extracted once per benchmark prompt and shared across all evaluated video generators; the extraction schema separates per-actor actions (\texttt{Action}), explicit temporal ordering or overlap (\texttt{Relations}), non-action scene and layout constraints (\texttt{IF}, reported as \textbf{Other}), and verification notes (\texttt{Review}), while excluding generic subject-count, facial-identity, and face-visibility criteria.

\begin{evalpromptbox}{Checklist Extraction Prompt}
\begin{Verbatim}
Extract an evaluation checklist faithfully from the source description. Output JSON only with fields Action, Relations, IF, and Review. Do not assess a video. Do not invent actions or unstated constraints.

Schema:
{"Action":[{"person":"@Image 1","actions":[{"id":"A1_1","text":"Chops vegetables on a cutting board.","evidence":"verbatim source substring"}]}],"Relations":[{"type":"overlap","groups":[["A1_1"],["A2_1"]],"evidence":"verbatim source substring"}],"IF":[{"text":"The scene is a kitchen.","evidence":"verbatim source substring"}],"Review":[]}

Action:
Include every @Image actor mentioned in the source once, in numerical order. Each observable action belongs only to its actor and appears once. Preserve the stated object, recipient, body part, direction, and manner qualifiers. Resolve pronouns to explicit nouns or @Image tags when unambiguous in the source; keep pronouns only when the referent appears in the same item. Evidence must be an exact nonempty contiguous substring from the source. Shared physical interactions may retain the partner tag (e.g., "lifts a crate with @Image 2") without inventing unstated reciprocal actions. Exclude person counts, facial identity, face visibility, and generic quality criteria.

Relations:
Record explicit temporal constraints once here instead of repeating timing clauses inside Action texts. Each group represents an interval covering its listed action IDs: "overlap" indicates overlapping intervals, and "before" indicates strict temporal precedence explicitly stated by markers such as "then", "before", or "after" (ordinary "and" or sentence order is insufficient). Keep one relation per explicit timing clause without pairwise expansion. Quote the exact timing clause in evidence. If an explicit timing clause is genuinely ambiguous, record {"evidence":"exact clause","reason":"specific uncertainty"} in Review instead of guessing; otherwise leave Review empty.

IF:
Extract explicitly stated, visually verifiable non-action requirements: scene setting, lighting, weather, camera movement, objects, object states, and static spatial arrangements not already entailed by Action items. Exclude human counts, extra-person bans, facial identity, and face visibility. Every IF item must cite an exact contiguous source substring.

Example:
Source: In a kitchen, @Image 1 chops vegetables on a cutting board while @Image 2 stirs a pot on the stove.
Output: {"Action":[{"person":"@Image 1","actions":[{"id":"A1_1","text":"Chops vegetables on a cutting board.","evidence":"@Image 1 chops vegetables on a cutting board"}]},{"person":"@Image 2","actions":[{"id":"A2_1","text":"Stirs a pot on the stove.","evidence":"@Image 2 stirs a pot on the stove"}]}],"Relations":[{"type":"overlap","groups":[["A1_1"],["A2_1"]],"evidence":"@Image 1 chops vegetables on a cutting board while @Image 2 stirs a pot on the stove"}],"IF":[{"text":"The scene is a kitchen.","evidence":"In a kitchen"}],"Review":[]}

Extraction Rules:
- Action granularity: Represent each coherent physical operation as a single item instead of splitting concurrent posture, grip, or expression modifiers into separate items; split only genuinely distinct sequential or parallel tasks.
- Interaction boundaries: Keep each actor's item focused on that actor's own behavior without duplicating a partner's action clause, and do not add unstated receiving actions when the text only names a recipient.
- Visual verifiability: Include only visually observable attributes and scene conditions; omit non-visual backstory, intent, or subjective attributes.
- Scene-vs-action separation: Place dynamic operations and intrinsic motion targets (such as walking toward a door or placing an item on a table) in Action, and place static actor-location or layout arrangements in IF without duplicating conditions already implied by an Action item. Do not infer unstated spatial assignments.
\end{Verbatim}
\end{evalpromptbox}

\begin{evalpromptbox}[promptTeal]{Action and Other Evaluation Prompt}
\begin{Verbatim}
Evaluate both Action items and IF (reported as Other) items independently using the same 16 chronological video frames (indexed 0 to 15). Role boxes R1, R2, ... denote tracked participants matched to @Image 1, @Image 2, ... and represent identities instead of screen positions. Extracted temporal Relations are provided as timing context for Action items. Scene fulfillment does not imply action fulfillment, and failed actions do not by themselves violate non-action scene requirements. Return a single JSON ratings array containing every supplied Action (A*_*) and IF (I*) item ID exactly once.

<ACTION_RUBRIC>
Judge whether the specified participant (Rk) performs the stated action, including its object, recipient, direction, manner, and explicit temporal relations. Actions performed by a different participant do not satisfy the item, and interaction items require all named partner tags to be verifiable. Evaluate across the 16-frame sequence: dynamic movements require observable change across frames, whereas static poses or holding states may be satisfied within relevant frames without requiring full-clip persistence unless explicitly stated.

First determine whether the frames provide sufficient visual evidence to judge the item. Return null when a required role tag is absent on relevant frames, or when occlusion or small spatial scale prevents distinguishing fulfillment from nonfulfillment. Do not convert visual uncertainty into intermediate scores (2-4).

For visually verifiable items, assign an integer score from 1 to 5:
1 = The target participant is adequately visible, but the requested action is not performed, contradicted, or directed to the wrong object or partner.
2 = Clear initial attempt toward the stated action with minimal task completion.
3 = Partial completion of the stated action, with a substantial requested component visibly unfinished.
4 = Main action completed, with only a minor visible shortfall in manner or qualifier.
5 = Stated action fully supported at the supplied visual resolution.

Scores 2-4 require a concrete, visible unfulfilled component of the stated requirement instead of a confidence discount. Do not penalize for cinematic style or unstated intermediate steps, and do not inflate scores merely because the surrounding scene looks plausible.
</ACTION_RUBRIC>

<IF_RUBRIC>
Judge each non-action requirement (scene setting, lighting, weather, camera movement, objects, object states, and static spatial layout) as written. Person-relative spatial items require the named Rk role tags, whereas global scene and object items do not. Participant counts and facial identity are evaluated by dedicated metrics and excluded here; do not re-score individual actions.

Assign an integer score from 1 to 5, or null when visual evidence is insufficient:
1 = The relevant scene, object, or layout is visible, but the stated requirement is contradicted or unmet.
2 = Only a minor part of the stated requirement is satisfied, with major discrepancies remaining.
3 = A meaningful part of the requirement is satisfied, while a substantial component remains unmet.
4 = Main requirement satisfied, with only a minor visible discrepancy.
5 = Stated requirement fully supported at the supplied visual resolution.
Return null when occlusion, missing role tags on person-relative layout items, or indistinguishable small details prevent verification. Deduct points only when visible evidence contradicts or leaves unmet the explicit wording of the item, not when unstated stereotypical fixtures are absent.
</IF_RUBRIC>

Output Format:
Return JSON only:
{"ratings":[{"id":"A1_1","score":5,"observed":"Concise visible evidence for the named role.","missing":"none, concrete unmet requirement, or reason for null","evidence_frames":[0,15]},{"id":"I1","score":5,"observed":"Concise visible evidence for the scene or layout item.","missing":"none","evidence_frames":[0,15]}]}
Include every supplied ID exactly once, cite valid frame indices (0-15) only, keep observed and missing fields consistent with score, and do not compute an overall score.
\end{Verbatim}
\end{evalpromptbox}

\twocolumn

%% file: tables/k_common_metrics.tex
\begin{table*}[!tp]
\centering
\small
\caption{Comparison on shared videos.}
\label{tab:k-common}
\begingroup
\setlength{\tabcolsep}{2.5pt}
\renewcommand{\arraystretch}{1.05}
\begin{tabular}{@{}lrrrrrrrrrrr@{}}
\toprule
\textbf{Method} & \textbf{$N$} & \textbf{Total} & \textbf{Face} & \textbf{IDCov} & \textbf{Count} & \textbf{TempID} & \textbf{Garment} & \textbf{IQ} & \textbf{Aes} & \textbf{Action} & \textbf{Other} \\
\midrule
Seedance-2.5~\citep{byteplus2026seedance25} & 229 & 75.04 & 0.582 & 0.956 & 0.939 & 0.726 & 0.720 & 64.91 & 5.26 & 16.36 & 8.79 \\
MiniMax-H3~\citep{minimax2026h3} & 229 & 73.45 & 0.380 & 0.948 & 0.984 & 0.599 & 0.781 & 74.84 & 5.51 & 17.73 & 9.45 \\
HappyHorse-1.1~\citep{alibaba2026happyhorse11} & 229 & 73.28 & 0.443 & 0.961 & 0.964 & 0.627 & 0.797 & 75.42 & 5.45 & 16.29 & 9.20 \\
Kling-4.0~\citep{kling2026v4} & 229 & 74.45 & 0.508 & 0.945 & 0.930 & 0.670 & 0.755 & 75.14 & 5.44 & 16.64 & 9.01 \\
Wan-3.0~\citep{alibaba2026wan3} & 229 & 74.37 & 0.511 & 0.972 & 0.953 & 0.683 & 0.836 & 74.59 & 5.34 & 15.96 & 8.66 \\
\midrule
BindWeave-14B~\citep{li2025bindweave} & 229 & 25.38 & 0.001 & 0.018 & 0.656 & 0.428 & 0.459 & 48.28 & 4.31 & 1.04 & 3.72 \\
LoomVideo-5B~\citep{wu2026loomvideo} & 229 & 39.38 & -0.001 & 0.056 & 0.823 & 0.544 & 0.656 & 66.25 & 5.02 & 5.65 & 6.98 \\
MAGREF-14B~\citep{deng2026magref} & 229 & 39.14 & 0.019 & 0.091 & 0.817 & 0.523 & 0.683 & 71.62 & 5.21 & 4.53 & 6.70 \\
OmniWeaving-8.3B~\citep{pan2026omniweaving} & 229 & 39.41 & 0.020 & 0.108 & 0.722 & 0.466 & 0.662 & 70.68 & 4.94 & 7.07 & 6.18 \\
VACE-14B~\citep{jiang2025vace} & 229 & 41.29 & 0.012 & 0.082 & 0.883 & 0.503 & 0.668 & 75.30 & 5.33 & 5.39 & 7.44 \\
Kaleido-14B~\citep{zhang2025kaleido} & 229 & 41.97 & 0.069 & 0.222 & 0.792 & 0.589 & 0.703 & 65.53 & 5.07 & 4.58 & 7.39 \\
RefAlign-1.3B~\citep{wang2026refalign} & 229 & 41.28 & 0.065 & 0.204 & 0.797 & 0.592 & 0.729 & 75.49 & 5.26 & 3.54 & 6.81 \\
RefAlign-14B~\citep{wang2026refalign} & 229 & 45.84 & 0.113 & 0.340 & 0.869 & 0.522 & 0.737 & 76.37 & 5.43 & 4.70 & 7.77 \\
UniVideo-13B~\citep{wei2025univideo} & 229 & 47.00 & 0.216 & 0.491 & 0.760 & 0.650 & 0.784 & 68.12 & 5.14 & 3.87 & 6.81 \\
VINO-13B~\citep{chen2026vino} & 229 & 47.80 & 0.209 & 0.471 & 0.790 & 0.631 & 0.737 & 70.17 & 5.21 & 4.97 & 6.74 \\
Phantom-14B~\citep{liu2025phantom} & 229 & 48.49 & 0.201 & 0.424 & 0.837 & 0.640 & 0.742 & 74.67 & 5.37 & 4.40 & 7.51 \\
MiniMax-H3-33B~\citep{minimax2026h3} & 229 & 65.22 & 0.434 & 0.907 & 0.917 & 0.577 & 0.724 & 73.93 & 5.38 & 11.42 & 8.17 \\
\midrule
\rowcolor{black!4}
\textbf{GroupRef-1.3B} & 229 & 47.72 & 0.195 & 0.493 & 0.830 & 0.614 & 0.833 & 71.45 & 5.30 & 3.88 & 6.95 \\
\rowcolor{black!4}
\textbf{GroupRef-14B} & 229 & 51.74 & 0.241 & 0.652 & 0.864 & 0.648 & 0.763 & 70.28 & 5.11 & 5.50 & 6.92 \\
\rowcolor{black!4}
\textbf{GroupRef-33B} & 229 & 65.78 & 0.464 & 0.920 & 0.895 & 0.571 & 0.725 & 73.66 & 5.35 & 11.46 & 8.31 \\
\bottomrule
\end{tabular}
\endgroup

\end{table*}

%% file: tables/action_coverage_all.tex
\begin{table*}[!tp]
\centering\small
\caption{Action coverage and execution.}
\label{tab:action-coverage-all}
\begin{tabular}{lrrrr}
\toprule
\textbf{Method} & \textbf{$N$} & \textbf{Coverage} & \textbf{Conditional} & \textbf{Action} \\
\midrule
\multicolumn{5}{l}{\textit{Commercial models}} \\
Seedance-2.5~\citep{byteplus2026seedance25} & 259 & 0.901 & 4.58 & 16.11 \\
MiniMax-H3~\citep{minimax2026h3} & 279 & 0.972 & 4.64 & 17.68 \\
HappyHorse-1.1~\citep{alibaba2026happyhorse11} & 280 & 0.950 & 4.39 & 16.12 \\
Kling-4.0~\citep{kling2026v4} & 320 & 0.903 & 4.62 & 16.34 \\
Wan-3.0~\citep{alibaba2026wan3} & 320 & 0.932 & 4.38 & 15.77 \\
\midrule
\multicolumn{5}{l}{\textit{Open-source baselines}} \\
BindWeave-14B~\citep{li2025bindweave} & 320 & 0.139 & 2.27 & 0.88 \\
LoomVideo-5B~\citep{wu2026loomvideo} & 320 & 0.644 & 2.78 & 5.74 \\
MAGREF-14B~\citep{deng2026magref} & 320 & 0.736 & 2.26 & 4.64 \\
OmniWeaving-8.3B~\citep{pan2026omniweaving} & 320 & 0.827 & 2.72 & 7.13 \\
VACE-14B~\citep{jiang2025vace} & 320 & 0.730 & 2.46 & 5.33 \\
Kaleido-14B~\citep{zhang2025kaleido} & 320 & 0.600 & 2.48 & 4.45 \\
RefAlign-1.3B~\citep{wang2026refalign} & 320 & 0.620 & 2.11 & 3.45 \\
RefAlign-14B~\citep{wang2026refalign} & 320 & 0.740 & 2.29 & 4.76 \\
UniVideo-13B~\citep{wei2025univideo} & 320 & 0.539 & 2.41 & 3.81 \\
VINO-13B~\citep{chen2026vino} & 320 & 0.607 & 2.59 & 4.82 \\
Phantom-14B~\citep{liu2025phantom} & 320 & 0.646 & 2.31 & 4.23 \\
MiniMax-H3-33B~\citep{minimax2026h3} & 320 & 0.871 & 3.55 & 11.10 \\
\midrule
\rowcolor{black!4}
\textbf{GroupRef-1.3B} & 320 & 0.726 & 2.08 & 3.93 \\
\rowcolor{black!4}
\textbf{GroupRef-14B} & 320 & 0.828 & 2.29 & 5.34 \\
\rowcolor{black!4}
\textbf{GroupRef-33B} & 320 & 0.897 & 3.50 & 11.23 \\
\bottomrule
\end{tabular}
\end{table*}